\documentclass[journal]{IEEEtran}
\pdfoutput=1

\usepackage[T1]{fontenc}
\usepackage{times}
\usepackage{hyperref}
\usepackage{url}
\usepackage{cite}
\usepackage{amsmath,amssymb,amsfonts}
\usepackage{algorithmicx}
\usepackage{graphicx}
\usepackage{textcomp}
\usepackage{xcolor}
\usepackage{color,soul}
\usepackage{balance}
\usepackage{flushend}
\usepackage{caption}
\usepackage{subcaption}
\usepackage{algorithm}
\usepackage[noend]{algpseudocode}
\usepackage{multirow}

\algnewcommand\algorithmicforeach{\textbf{for each}}
\algdef{S}[FOR]{ForEach}[1]{\algorithmicforeach\ #1\ \algorithmicdo}

\begin{document}


\title{
 Hyperseed: Unsupervised Learning with Vector Symbolic Architectures
 }

\author{Evgeny~Osipov, 
Sachin~Kahawala, Dilantha~Haputhanthri, Thimal~Kempitiya, Daswin~De~Silva, Damminda~Alahakoon, Denis~Kleyko

\thanks{
Manuscript received on October 19, 2021; revised on June 30, 2022; accepted on September 18, 2022.
This work was supported in part by the Intel Neuromorphic Research Community Grant to Lule\aa~ University of Technology, the Swedish Foundation for international Cooperation in Research and Higher Education (STINT) under Mobility Grant for Internationalisation MG2020-8842, and the Russian Science Foundation during the period of 2020-2021 under grant 20-71-10116.
SK and DH are recipients of the Centre for Data Analytics and Cognition (CDAC) Ph.D. research scholarships. 
DK has received funding from the European Union's Horizon 2020 research and innovation programme under the Marie Skłodowska-Curie grant agreement No 839179.
}
\thanks{E. Osipov is with the Department of Computer Science, Electrical and Space Engineering at Luleå University of Technology, 97187 Luleå, Sweden.  \mbox{E-mail}: \mbox{Evgeny.Osipov@ltu.se}
}
\thanks{S. Kahawala, D. Haputhanthri, T.~Kempitiya, D.~De~Silva and D.~Alahakoon  are with the Centre for Data Analytics and Cognition (CDAC) at La Trobe University, Melbourne, Australia.   \mbox{E-mail}: \{S.Kahawala, D.Haputhanthri, T.Kempitiya,
 D.DeSilva, D.Alahakoon\}  @latrobe.edu.au
 }

\thanks{D. Kleyko is with the Redwood Center for Theoretical Neuroscience at the University of California, Berkeley, CA 94720, USA and also with the Intelligent Systems Lab at Research Institutes of Sweden, 16440 Kista, Sweden. \mbox{E-mail}: \mbox{denis.kleyko@ri.se}
}

}
\newcommand{\fix}{\marginpar{FIX}}
\newcommand{\new}{\marginpar{NEW}}

\maketitle

\begin{abstract} 

Motivated by recent innovations in biologically-inspired neuromorphic hardware, this article presents a novel unsupervised machine learning algorithm named Hyperseed that draws on the principles of Vector Symbolic Architectures (VSA) for fast learning of a topology preserving feature map of unlabelled data.  It relies on two major operations of VSA, binding and bundling. The algorithmic part of Hyperseed is expressed within Fourier Holographic Reduced Representations model, which is specifically suited for implementation on spiking neuromorphic hardware.  The two primary contributions of the Hyperseed algorithm are, few-shot learning and a learning rule based on single vector operation. These properties are empirically evaluated on synthetic datasets as well as on illustrative benchmark use-cases, IRIS classification, and a language identification task using $n$-gram statistics. The results of these experiments confirm the capabilities of Hyperseed and its applications in neuromorphic hardware. 
\end{abstract}


\section{Introduction}
\label{sect:intro}
Vector Symbolic Architectures (VSA) are increasingly leveraged and adapted in machine learning and robotics algorithms and applications \cite{HDGestureIEEE, Superposition_Olshausen, NeubertRobotics2019, HerscheCompressingBCI2020, Kleyko_RVFL, NeubertAggregation2021}. In classification tasks, the use of VSA leads to order of magnitude increase in energy efficiency of computations on the one hand and natively enables one-shot and multi-task learning on the other~\cite{ChangHDTaskProjected2020, ChangHDInformationPreserved2020, KarunaratneInMemory2020,KarunaratneHDAugmented2021, KleykoAugmented2022}. It is prospected that VSA will play a key role in the development of novel neuromorphic computer architectures~\cite{Loihi18} as an algorithmic abstraction \cite{RahimiNanoscalable2017, KleykoComputingParadigm2021}. The main contribution of this article is a novel algorithm for unsupervised learning called Hyperseed, which relies on the mathematical properties arising from random high-dimensional representation spaces described through the phenomenon of the concentration of measure \cite{Gorban2018Blessing} and of the main VSA operations of binding and superposition~\cite{Kanerva09}. The method's name suggests that data samples are encoded as high-dimensional vectors (also called \textit{hypervectors}, HVs or ``seeds''), which are then mapped (a.k.a. ``sowed'') onto a specially prepared topologically arranged set of hypervectors for revealing the internal cluster structure in the unlabeled data.

The Hyperseed algorithm bears conceptual similarities to Kohonen's Self-Organizing Maps (SOM) algorithm \cite{SOMBook, intSOM}, therefore, selected SOM terminology is adopted for the description of our approach. 
However, it is designed using cardinally different computing principles to the SOM algorithm. 
Hyperseed implements the entire learning pipeline in terms of VSA operations. To the best of our knowledge, this has not been attempted prior and is reported for the first time. 

The Hyperseed algorithm is presented using Frequency Holographic Reduced Representations (FHRR)~\cite{PlateNested1994} model of VSAs and the concept of Fractional Power Encoding (FPE) \cite{PlateNested1994, frady2021computing, FradyFunctionsNICE2022, Komer2019ANR, komer2020biologically}.  The usage of the FHRR model makes the proposed solution specifically fit for implementation on spiking neural network architectures including Intel's Loihi \cite{Loihi18}.

To this end, we introduce the Hyperseed algorithm and demonstrate its performance on three illustrative non-linear classification problems of varying complexity: Synthetic datasets from  the Fundamental Clustering Problems  Suite, Iris classification, and language identification using $n$-gram statistics. Across all experiments, Hyperseed convincingly demonstrates its key novelties of learning from a few input vectors and single vector operation learning rule, both of which contribute towards reduced time and computation complexity. 

The article is structured as follows.  
Section~\ref{sect:related} describes related work to Hyperseed operations. The VSA methods leveraged in Hyperseed are presented in Section \ref{sect:method}.
Section~\ref{sect:vsaseed} presents the main contribution -- the method for unsupervised learning -- Hyperseed.  
Section~\ref{sect:perf} reports the results of the experimental performance evaluation.  
Section~\ref{sect:discussion} discusses the suitability of Hyperseed for realization on neuromorphic hardware.  
The conclusions follow in Section~\ref{sect:conclusions} .

\section{Related Work}
\label{sect:related}
 
VSA  \cite{PlateNested1994, Rachkovskij2001, KleykoSDR2016,  FradySDR2020} is a computing framework providing methods of representing and manipulating concepts and their meanings in a high-dimensional space. VSA finds its applications in, for example, 
cognitive architectures~\cite{BuildBrain,RachkovskijAnalogical2004,RachkovskijAnalogy2012}, 
natural language processing~\cite{BICA16CT,JonesMeaning2007, RPRSKJ2015, RachkovskijRecursiveBinding2022, RachkovskijEquivariant2021}, communications~\cite{JakimovskiCollective2012, KleykoMACOM2012, KimHDM2018},
biomedical signal processing~\cite{ACCESS_HRV, HDGestureIEEE}, approximation of conventional data structures~\cite{HD_FSA, ABF},
and for classification tasks such as gesture recognition~\cite{TNNLS18, HDGestureIEEE}, cybersecurity threat detection \cite{christopher2021minority, moraliyage2022evaluating}, physical activity recognition~\cite{Rasanen14}, character recognition~\cite{goltsev2005combination, Rachkovskij2022NCA}, speaker identification~\cite{HuangSpeaker2022}, fault isolation and diagnostics~\cite{KussulDiagnostics1998, ACCESS_BIOFAULT, EggimannConfigurableHD2021}. Examples of efforts on using VSA for other than classification learning tasks are using data HVs for clustering~\cite{ImaniHDCluster2019, BandaragodaTrajectoryTraffic2019, HernandezClustering2021}, semi-supervised learning~\cite{ImaniSemiHD2019}, collaborative privacy-preserving learning~\cite{ImaniHDColLearn2019, KhaleghiPriveHD2020}, multi-task learning~\cite{ChangHDTaskProjected2020, ChangHDInformationPreserved2020}, distributed learning~\cite{RosatoHDDistributed2021, HsiehFL2021}.
A comprehensive two-part survey of VSA is available in~\cite{KleykoSurveyVSA2021Part1,KleykoSurveyVSA2021Part2}. 

Hypervectors of high (but fixed) dimensionality (denoted as $d$) are the basis for representing information in VSA.  
The information is distributed across hypervector positions, therefore, hypervectors use distributed representations \cite{Hinton1986}. There are different VSA models that all offer the same operation primitives but differ slightly in terms of the implementation of these primitives. For example, there are VSA models that compute with binary, bipolar \cite{Kanerva:Hyper_dym13, MAP}, continuous real, and continuous complex vectors \cite{PlateNested1994}. Thus, the VSA concept has the flexibility to connect to a multitude of different hardware types, such as binary-valued VSAs for analog in-memory computing architectures~\cite{KarunaratneInMemory2020} or complex-valued VSAs for spiking neuron architectures~\cite{TPAM, RennerBinding2022, BentSpike2022}.

The relevant sub-domain of related work to the proposed Hyperseed algorithm  is the application of VSA for solving machine learning tasks. In this context, VSA have been used for: 1) Representing input data and interfacing such representations with conventional machine learning algorithms and 2) Implementing the functionality of neural networks with VSA operations. 

The most illustrative use cases for encoding of input data into hypervectors and interfacing conventional machine learning algorithms are \cite{BandaragodaTrajectoryTraffic2019, RachkovskijClassifiers2007, RIJHK2015, AlonsoHyperEmbed2020, MirusBehavior2019, KleykoBoostingSOM2019, MirusBalanced2020, ShridharEnd2End2020, Kussul1999IJCNN, Rachkovskij2015Cybern}. For example, works~\cite{AlonsoHyperEmbed2020, KleykoBoostingSOM2019} proposed encoding $n$-gram statistics into hypervectors and subsequently solving typical natural language processing tasks with either supervised or unsupervised learning using standard artificial  neural network architectures. The main distinctive property of VSA represented data is the substantial reduction of the memory footprint and the reduced learning time.  In \cite{BandaragodaTrajectoryTraffic2019}, hypervectors were used to encode sequences of variable lengths in the context of unsupervised learning of traffic patterns in intelligent transportation system application.
In the context of visual navigation, hypervectors were used as input to Simultaneous Localization and Mapping (SLAM) algorithms \cite{NeubertRobotics2019} as well as for ego-motion estimation~\cite{MitrokhinSensorimotor2019, KleykoCommentariesSR2020, HerscheDVSCDT2020}.

A great potential of VSAs was demonstrated when used for the implementation of the entire functionality of some classical neural network architectures. In~\cite{Kleyko_RVFL, KleykointESN2020, DiaoGLVQHD2021} the functionality of an entire class of randomly connected neural networks (random vector functional link networks~\cite{IgelnikRVFL1995} and echo state networks~\cite{RC09}) was implemented purely in terms of VSA operations.
It was demonstrated that implementing the algorithm functionality with bipolar VSAs allows reducing energy consumption on the specialized digital hardware by the order of magnitude, while substantially decreasing the operation times. 
Moreover, further flexibility can be achieved~\cite{KleykoCA2020, EggimannConfigurableHD2021} when considering the ways of generating random connections used in the networks.

The main contribution of this article in the context of VSA is a novel approach to learning since the dominating learning approach in the area is on creating a single hypervector for a specific class.
This is achieved through encoding input data and then forming associative memory storing the prototypical representations for individual classes. 
Our approach to learning is radically different -- it utilizes the similarity preservation property of the binding operation in combination with the FPE encoding method~\cite{PlateNested1994}. 
FPE was recently used to simulate and predict dynamical systems~\cite{VoelkerFPEDynamical2021}, perform integer factorization~\cite{KleykoPrimes2022}, and represent order in time series~\cite{SchlegelHDC-MiniROCKET2022}.
The associative memory in the proposed approach is created once during the initialization phase and remains fixed during the life-time of the system. The update requires a single vector operation. To the best of our knowledge, this is the first reseacrh article to present the usage of VSA in unsupervised learning tasks.

\section{Method: Holographic Reduced Representations (HRR) model}
\label{sect:method}

The Hyperseed algorithm is designed using the Fourier Holographic Reduced Representations (FHRR)  model \cite{PlateNested1994}. 
FHRR facilitates the mathematical treatment of Hyperseed operations. The potential argument that complex numbers used in FHRR add to the memory requirements of Hyperseed is intuitively true in the case of CPU realization. However, in Section \ref{sect:discussion}, we rationalise this is not an issue for the neuromorphic hardware. Also, due to the equivalence of FHRR and HRR models, the operations of Hyperseed can be implemented with hypervectors from $\mathbb{R}^d$.  
In fact, when evaluating the performance of the bottleneck functionality of Hyperseed on Intel's Loihi, we use HRR model\footnote{
The supplementary code base also contains the HRR implementation of the algorithm.
}. 
The atomic FHRR hypervectors are randomly sampled from $\mathbb{C}^d$.  Dimensionality $d$ is a hyperparameter of Hyperseed. In high-dimensional random spaces, all random hypervectors are dissimilar to each other (quasi-orthogonal) with an extremely high probability. VSA defines operations and a similarity measure  on hypervectors. In this article, we use the cosine similarity of real parts of hypervectors for characterizing the similarity.
The three primary operations for computing with hypervectors are superposition, binding, and permutation. 

\subsection{Binding operation} 
The binding operation is used to bind two hypervectors together. The result of the binding is another hypervector. For example, for two hypervectors $\textbf{v}_1$ and $\textbf{v}_2$  the result of binding of their hypervectors (denoted as $\textbf{b}$) is calculated as follows: 
%
\begin{equation}
\label{eq:bind} 
\textbf{b} = \textbf{v}_1 \circ \textbf{v}_2,  
\end{equation}
where the notation $\circ$  is used to denote the binding operation. 
In HRR, the binding operation is implemented as circular convolution of $\textbf{v}_1$  and $\textbf{v}_2$, which can be implemented as the component-wise multiplication in the Fourier domain. This observation inspired FHRR where the representations are already in the Fourier domain in a form of phasors so that the component-wise multiplication, which is equivalent to the addition of phase angles modulo $2\pi$, plays the role of the binding operation.  
Binding is, essentially, a randomizing operation that moves hypervectors to another (random) part of the high-dimensional space. 
The role played by the binding operation depends on the algorithmic context. 
In data structures with roll-filler pairs, the binding operation corresponds to the assignment of a value (filler) to a variable (role). There are two important properties of the binding operation. First, the resultant hypervector $\textbf{b}$ is dissimilar to the hypervectors being bound, i.e., the similarity between $\textbf{b}$ and $\textbf{v}_1$ or $\textbf{v}_2$ is approximately $0$.

Second, the binding operation preserves similarity. That is the distribution of the similarity measure between hypervectors from some set $\mathcal{S}$ is preserved after binding of all hypervectors in $\mathcal{S}$ with the same random hypervector $\textbf{v}$.  

The binding operation is reversible. The unbinding, denoted as $\oslash$,  is implemented by the circular correlation in HRR.  In the case of FHHR this is equivalent to component-wise multiplication with the complex conjugate. Being the inverse of the binding operation, the unbinding obviously has the same similarity preservation property when performed on all hypervectors in $\mathcal{S}$ with the same  hypervector $\textbf{v}$:   

 \begin{equation}
\label{eq:unbind} 
\textbf{v}_2 \oslash \textbf{b}  =  \textbf{v}_1.  
\end{equation}
The interpretation of the unbinding operation is a retrieval of a value from the hypervector encoding the assignment. When unbinding is performed from the superposition of bindings (see Section~\ref{sec:vsa:superposition}), the retrieved hypervector contains noise. In VSA, the noisy vector can be cleaned-up by performing a search for the closest atomic hypervector stored in an associative memory.

\subsection{Permutation operation}
The permutation (rotation) operation $\textbf{b} = \rho(\textbf{v})$ is a unitary operation that is commonly used to represent an order of the symbol in a sequence.  As with the binding operation, the resultant hypervector $\textbf{b}$ is dissimilar to $\textbf{v}$. In this article, this operation is used for encoding a certain type of input data as further described in Section \ref{sect:perf}.

\subsection{Superposition operation}
\label{sec:vsa:superposition}
Superposition is denoted with $+$ and implemented via component-wise addition. 
The superposition operation combines several hypervectors into a single hypervector. 
For example, for hypervectors $\textbf{v}_1$ and $\textbf{v}_2$  the result of superposition (denoted as $\textbf{a}$) is simply: 
%
\begin{equation}
\label{eq:bindle} 
\textbf{a} = \textbf{v}_1 + \textbf{v}_2.
\end{equation}
In contrast to the binding operation, the resultant hypervector $\textbf{a}$ is similar to all superimposed hypervectors, i.e., the cosine similarity between $\textbf{b}$ and $\textbf{v}_1$ or $\textbf{v}_2$ is larger than $0$.
If several copies of any hypervector are included (e.g., $\textbf{a} = 3\textbf{v}_1 + \textbf{v}_2$), the resultant hypervector is more similar to the dominating hypervector than to other components.  

If superposition is applied to several bindings it is possible to unbind any hypervector from any binding. In this case, the result of the unbinding operation is a noisy version of the second operand of the particular binding. For example, if $\textbf{a}=\textbf{v}_1\circ \textbf{v}_2 + \textbf{u}_1 \circ \textbf{u}_2$, then $\textbf{u}_2 \oslash \textbf{a}=\textbf{u}_1+\mathrm{noise}= \textbf{u}_1^*$. Given that noiseless atomic hypervectors ($\textbf{v}_1,\textbf{v}_2, \textbf{u}_1, \textbf{u}_2$) are kept in the associative memory and so vector $\textbf{u}_1^*$ is expected to have the highest  similarity to $\textbf{u}_1$.  The same property holds for the binding of any atomic hypervector with the superposition of unbindings (which we use below in the description of our approach). That is if $\textbf{a}=\textbf{v}_1\oslash \textbf{v}_2 + \textbf{u}_1 \oslash \textbf{u}_2$, then $\textbf{u}_2 \circ \textbf{a}=\textbf{u}_1+\mathrm{noise}= \textbf{u}_1^*$.

\section{Hyperseed: Unsupervised Learning with Vector Symbolic Architectures}
\label{sect:vsaseed}

This section presents the main   contribution of this article -- the method for unsupervised learning -- Hyperseed. Denote the set of FHRR-represented input data as $\mathcal{D}\in\mathbb{C}^d$. 
%
Data hypervectors for training are generated during the encoding phase (see Section \ref{sect:encoding} for the details of the encoding). They are kept in a working memory.  The input query for testing is also a data hypervector obtained using the same encoding procedure as for the training data. 

Denote as $\mathcal{P}\in\mathbb{C}^d$ the 
vector space with known similarity properties.   Set $\mathcal{P}$ is created by encoding points (also referred to as nodes further in the text) of a 2D grid using FPE method~\cite{PlateNested1994,frady2021computing, FradyFunctionsNICE2022, Komer2019ANR, komer2020biologically}. The cardinality of HD-map  $|\mathcal{P}|=n\times m$, where $n$ and $m$ are the sizes of the grid along the vertical and the horizontal axes, respectively. For the sake of brevity, further in the text we will refer to set   $\mathcal{P}$ as HD-map. HD-map is computed, as described below, once and is stored in the associative memory. This memory is fixed throughout the life-time of the system.  

The Hyperseed algorithm relies on the similarity preservation property of the (un)binding operation.  The goal with Hyperseed is to translate the original data hypervectors $\mathcal{D}$ (with unknown) internal similarity layout to HD-map $\mathcal{P}$ by unbinding all of its members from hypervector $\mathbf{s}$, i.e.:

\begin{equation}
    \mathcal{D}\oslash \mathbf{s}\Rightarrow\mathcal{P}.
\label{eq:transf}
\end{equation}
\noindent
Hypervector $\mathbf{s}$ is obtained as the result of applying an unsupervised learning rule. In essence, during the learning, some selected hypervectors from $\mathcal{D}$ will be bound to selected vectors from $\mathcal{P}$ as described further\footnote{Due to an analogy of ``seeding'' data hypervectors onto HD-map, hypervector $\mathbf{s}$ is referred as seed vector throughout the article.}. 

\subsection{Initialization Phase: Generation of  HD-map $\mathcal{P}$ and hypervector $\mathbf{s}$}
\label{sect:hdMap}
The hypervectors, which are the members of $\mathcal{P}$, are generated such that the similarity between them relates to topological proximity of grid nodes. Note, however, that the reference to the topological arrangement of $\mathcal{P}$ is virtual in a sense that in the associative memory, in which hypervectors of $\mathcal{P}$ are stored does not have any structure. Topology information is kept on a side to be used for visualization purposes only.

\begin{figure}[t!]
\centerline{\includegraphics[width=0.7\columnwidth]{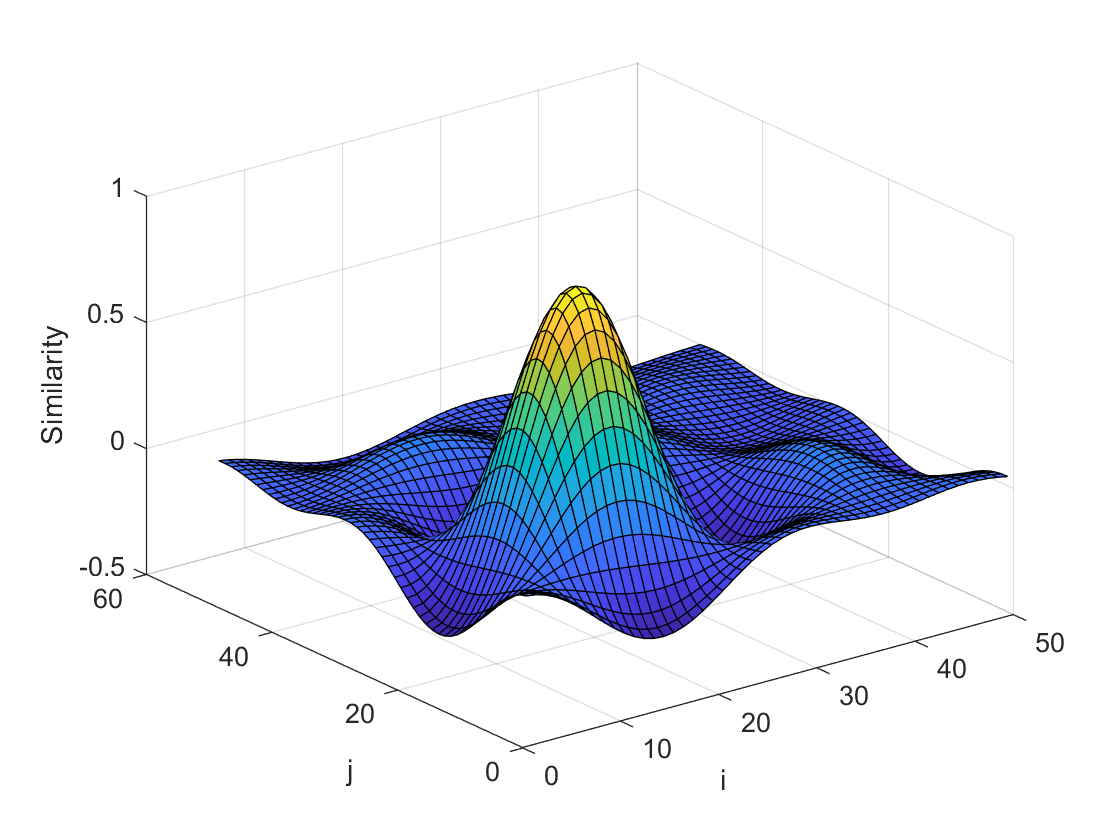}}
\caption{Similarity distribution on an HD-map. The target node is $(15,15)$, the size of the grid is $50\times50$, bandwidth $\epsilon=0.05$.
}
\label{fig:hdplane}
\end{figure}

The generation of HD-map starts with two randomly generated unit hypervectors $\mathbf{x}_0, \mathbf{y}_0 \in \mathbb{C}^d$ as $\mathbf{x}_0 \sim e^{j\cdot 2\pi\cdot U(0,1)}$ and $\mathbf{y}_0 \sim e^{j\cdot 2\pi\cdot U(0,1)}$.
Let us denote the bandwidth parameter regulating the similarity between the adjacent coordinates on the grid by $\epsilon$. The \textit{i}-th $x$ and $y$ coordinates of the grid will be created using the FPE method as:
\noindent
\begin{equation}
    \mathbf{x}_i=\mathbf{x}_0^{\epsilon \cdot i}, 
    \mathbf{y}_i=\mathbf{y}_0^{\epsilon \cdot i}.
    \label{eq:x_y}
\end{equation}
\noindent
The hypervector $\mathbf{p}_{(i,j)}$ representing a node with coordinates $(i,j)$ on the grid is computed as  $\mathbf{p}_{(i,j)}=\mathbf{x}_i \circ \mathbf{y}_j$. 
Fig.~\ref{fig:hdplane} illustrates the landscape of similarity between all hypervectors of HD-map stored in the associative memory and one selected hypervector of the same HD-map encoding coordinates (15,15) on $50\times50$ 2D grid as a function of coordinates $i$ and $j$.

Hypervector $\mathbf{s}$ is also initialized randomly: $\mathbf{s}  \sim e^{j\cdot 2\pi\cdot U(0,1)}$. It is updated over several iterations during the learning phase as described below. The number of update iterations is a hyperparameter of Hyperseed.  

\subsection{Search procedure in Hyperseed: Finding Best Matching Vector on HD-map}
\label{sect:search}
In the Hyperseed algorithm, HD-map $\mathcal{P}$ acts as the auto-associative memory~\cite{FrolovWillshaw2002,FrolovTime2006, GritsenkoAMSurvey2017}. That is the only operation performed on HD-map is the search for the Best Matching Vector (BMV) given some input hypervector. 
The BMV is found by computing the cosine similarity between the input hypervector and all hypervectors in $\mathcal{P}$. The output of this procedure is a valid hypervector in $\mathcal{P}$ with the highest similarity to the input hypervector. 

The mapping ($\mathbf{d}_i \rightarrow \mathbf{p}_i$) of data hypervectors in $\mathcal{D}$ to hypervectors of  HD-map $\mathcal{P}$  is done by unbinding $\mathbf{d}_i$ from the trained hypervector $\mathbf{s}$: 

 \begin{equation}
     \mathbf{p}_i^*=\mathbf{d}_i \oslash \mathbf{s}.
     \label{eq:seedbind}
 \end{equation}
 In (\ref{eq:seedbind}), $\mathbf{p}_i^*$ is a noisy version of a hypervector in $\mathcal{P}$.

\subsection{Update phase: Unsupervised learning of hypervector $\mathbf{s}$}
\label{sect:update}
 The goal with the update procedure on each iteration is to map input hypervector $\mathbf{d}_i$ as near as possible to some target hypervector in $\mathcal{P}$ with respect to the cosine similarity.
 
 Therefore, a single learning  iteration consists of three steps: 
 \begin{enumerate}
     \item Choose a target hypervector $\mathbf{p}_{\texttt{target}}$ (see the next subsection); 
     \item Compute a hypervector for the perfect mapping 
     $\mathbf{d}_i \rightarrow \mathbf{p}_{\texttt{target}}$ by binding of $\mathbf{d}_i$ with $\mathbf{p}_{\texttt{target}}$. 
     \item Update hypervector $\mathbf{s}$ by adding this perfect mapping hypervector to hypervector $\mathbf{s}$:

     \begin{equation}
     \mathbf{s}=\mathbf{s} +  \mathbf{d}_i \circ \mathbf{p}_{\texttt{target}}.
     \label{eq:seedupdate}
 \end{equation}
 \end{enumerate}
 \noindent
 Note that after the update, $\mathbf{s}$ is not a phasor vector anymore so it might be renormalized if necessary. 
 Thus, by the end of the learning phase, hypervector $\mathbf{s}$ is the superposition of bindings $\mathbf{d}_i \circ \mathbf{p}_j$. As such, the result of unbinding of hypervectors similar to $\mathbf{d}_i$ with hypervector $\mathbf{s}$ (\ref{eq:seedbind}) will resemble hypervectors in $\mathcal{P}$ (i.e., the hypervectors used in the update phase).
 
\subsection{Weakest match search (WMS) phase: Finding a data hypervector for the update in a single iteration}

\label{sect:observe}

To find a data hypervector for the update of hypervector $\mathbf{s}$ (\ref{eq:seedupdate}), Hyperseed uses a heuristic based on the  farthest-first traversal rule (FFTR). This principle is widely used for defining heuristics in many important computing applications ranging from approximation of Traveling Salesman Problem \cite{TravelingSalesman} to $k$-center clustering \cite{k-center} and fast similarity search \cite{Rachkovskij2017Cybern}. FFTR has been also used as a weight update rule in SOMs \cite{FFTSOM}, which resulted in better representation of outliers as well as lower topographic and quantization errors. In FFTR, the first point is selected arbitrarily and each successive point is as far as possible from the set of previously selected points.  In the case of Hyperseed, FFTR is straightforwardly implemented by checking the cosine similarity between the noisy vector $\mathbf{p}^*$ (\ref{eq:seedbind}) with all noiseless hypervectors of HD-map $\mathcal{P}$.  In order to demonstrate this, we shall consider the properties of the transformation from $\mathcal{D}$ to $\mathcal{P}$ with the unbinding operation (\ref{eq:transf}).

\begin{figure}[t!]
\centering
\includegraphics[width=6.5cm]{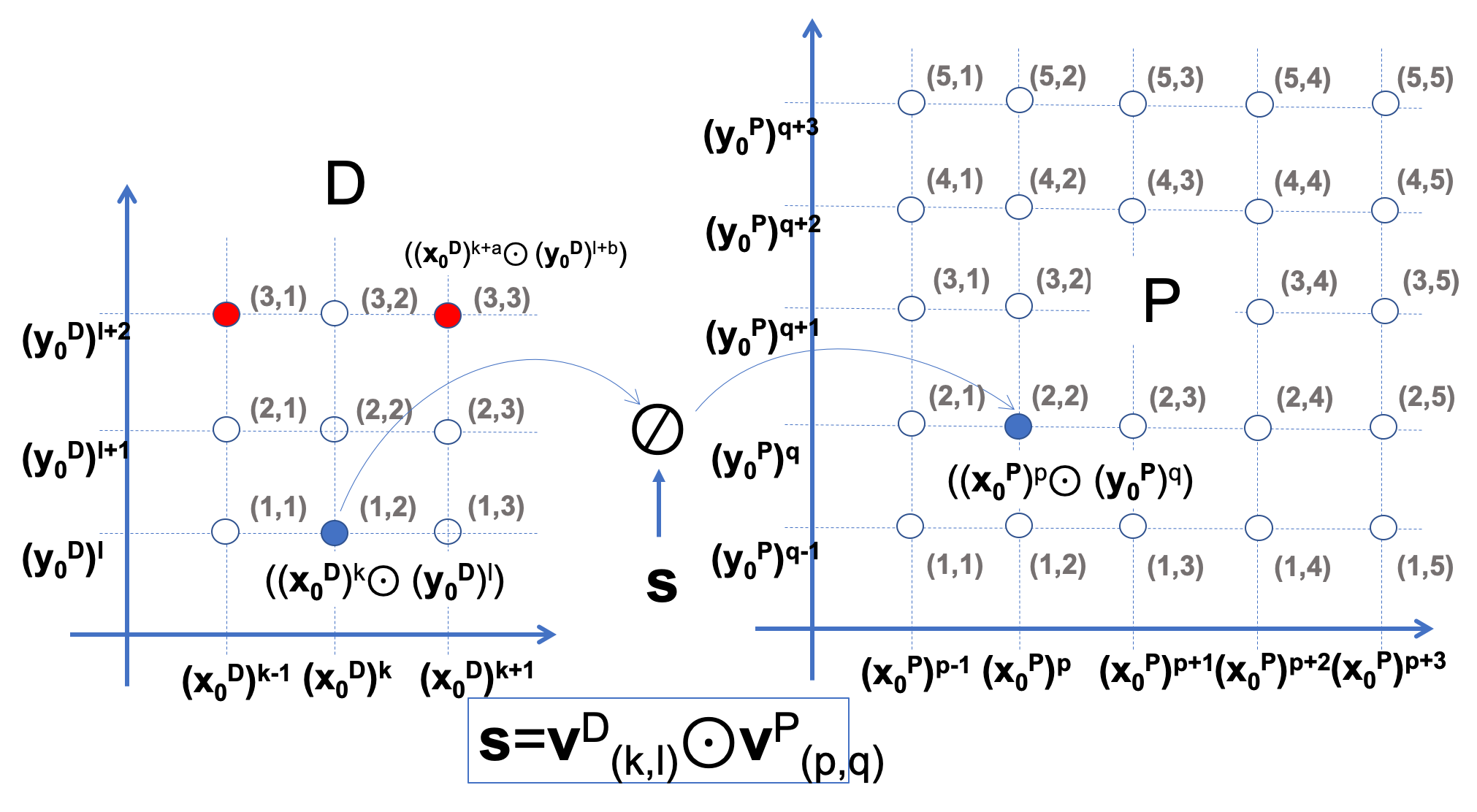}
\caption{Transformation of space $\mathcal{D}$ into $\mathcal{P}$ with the unbinding operation.
}
\label{fig:gridmapping}

\end{figure}

Consider an example of transforming vector space $\mathcal{D}$ of FPE-encoded two dimensional grid structure to HD-map  $\mathcal{P}$.  Graphically, the scenario is illustrated in Fig. \ref{fig:gridmapping}. The data is transformed via FPE as described in Section~\ref{sect:hdMap} with initial random bases $\mathbf{x}^\mathcal{D}_0$ and $\mathbf{y}^\mathcal{D}_0$. Importantly, while the transformed hypervectors in $\mathcal{D}$ are available for observation, the base hypervectors, which were used in the transformation are unknown. The bandwidth $\epsilon_{\mathcal{D}}$ used during the encoding is also unknown. 

Vector space $\mathcal{P}$ is represented via FPE, now with initial random bases $\mathbf{x}^P_0$ and $\mathbf{y}^P_0$. Importantly, the base hypervectors of $\mathcal{P}$ as well as FPE bandwidth $\epsilon_{\mathcal{P}}$ are known. For brevity of calculations, it is assumed the FPE bandwidth of vector space $\mathcal{D}$ is the same as that of vector space $\mathcal{P}$, that is $\epsilon_{\mathcal{P}}=\epsilon_{\mathcal{D}}=\epsilon$.

We now pick an arbitrary vector from $\mathcal{D}$ encoding a pair of values $(k,l)$ and bind it to a hypervector from $\mathcal{P}$ encoding some predefined pair of values $(p,q)$:\footnote{
Strictly speaking, equations below should include modulo $2\pi$ operations but they are omitted for the sake of readability. 
}  
\begin{equation}
\begin{split}
    \mathbf{s} &=e^{j\cdot 2\pi\epsilon_{\mathcal{P}}\cdot x^P_0\cdot p+j\cdot 2\pi\epsilon_{\mathcal{P}}\cdot y^P_0\cdot q +j\cdot 2\pi\epsilon_{\mathcal{D}}\cdot x^D_0\cdot k+j\cdot 2\pi\epsilon_{\mathcal{D}} \cdot y^D_0\cdot l}\\
    &=e^{j\cdot 2\pi \epsilon_{\mathcal{P}} \cdot (x^P_0\cdot p+y^P_0\cdot q+\frac{\epsilon_{\mathcal{D}}}{\epsilon_{\mathcal{P}}}x^D_0\cdot k+\frac{\epsilon_{\mathcal{D}}}{\epsilon_{\mathcal{P}}}y^D_0\cdot l)}. 
\end{split}
\label{eq:seedvector}
\end{equation}

Obviously, as the result of unbinding of the hypervector representing the coordinate $(k,l)$  from $\mathcal{D}$ from hypervector $\mathbf{s}$, it will be translated to the hypervector for $(p,q)$  from $\mathcal{P}$ as illustrated in Fig. \ref{fig:gridmapping}.

Let us now unbind a hypervector from $\mathcal{D}$ encoding a point on a certain offset $(a,b)$ from $(k,l)$, that is hypervector: $e^{j\cdot 2\pi\epsilon_{\mathcal{D}} \cdot (x^D_0\cdot (k+a) + y^D_0\cdot (l+b))}$, which is the farthest from point$(k,l)$ in this scenario. Recall that unbinding in FHRR is implemented as a component-wise multiplication with the complex conjugate:
\noindent
\begin{equation}
\begin{split}
    \mathbf{v}^* &=e^{j\cdot 2\pi \epsilon_{\mathcal{P}} \cdot (x^P_0\cdot p+y^P_0\cdot q+\frac{\epsilon_{\mathcal{D}}}{\epsilon_{\mathcal{P}}}x^D_0\cdot k+\frac{\epsilon_{\mathcal{D}}}{\epsilon_{\mathcal{P}}}y^D_0\cdot l)} \cdot \\
    &\cdot e^{j\cdot 2\pi\epsilon_{\mathcal{D}} \cdot (-x^D_0\cdot (k+a) - y^D_0\cdot (l+b))}\\
    &=e^{j\cdot 2\pi \epsilon_{\mathcal{P}} \cdot(x^P_0\cdot (p+\alpha_1\cdot a)+y^P_0\cdot (q + \alpha_2 \cdot b)}. 
\end{split}
\label{eq:unbindoffset}
\end{equation}
\noindent
In \ref{eq:unbindoffset}, $\alpha_1=-\frac{\epsilon_{\mathcal{D}} \cdot x^D_0}{\epsilon_{\mathcal{P}} \cdot x^P_0}$ and $\alpha_2=-\frac{\epsilon_{\mathcal{D}} \cdot y^D_0}{\epsilon_{\mathcal{P}} \cdot y^P_0}$ are coefficients introduced in order to align the result of unbinding with HD-map $\mathcal{P}$ for ease of the interpretation. 

\begin{figure}[t!]
        \centering
        \begin{subfigure}[b]{0.3\columnwidth}
            \centering
             \includegraphics[width=\textwidth]{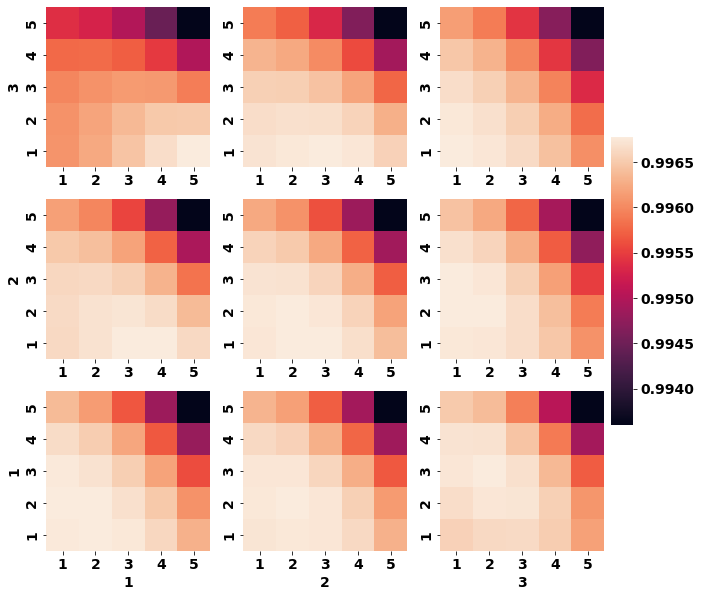}
            \caption[]%
            {{\small Similarities to $\mathbf{v}^P_{(i,j)}$,$\epsilon_{\mathcal{P}}<\epsilon_{\mathcal{D}}$.}}    
            \label{fig:fft1}
        \end{subfigure}
        \hfill
        \begin{subfigure}[b]{0.3\columnwidth}  
            \centering 
            \includegraphics[width=\textwidth]{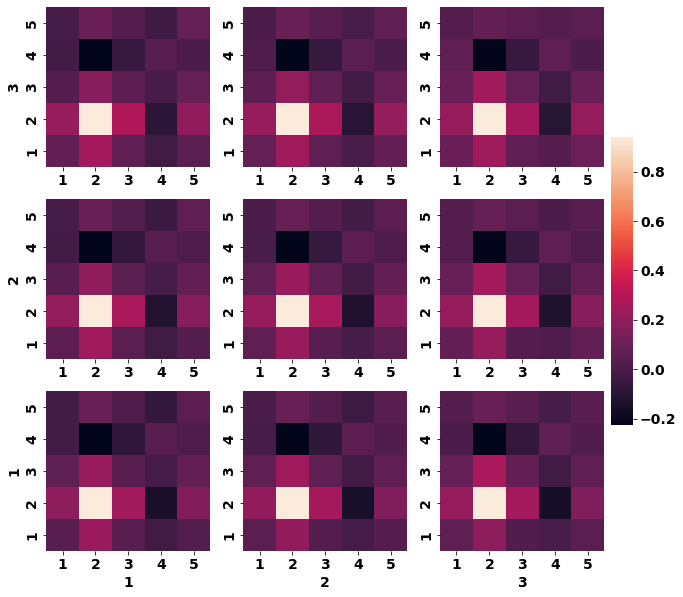}
            \caption[]%
            {{\small Similarities to $\mathbf{v}^P_{(i,j)}$, $\epsilon_{\mathcal{P}}>\epsilon_{\mathcal{D}}$.}}    
            \label{fig:fft2}
        \end{subfigure}
        \hfill
        \begin{subfigure}[b]{0.3\columnwidth}  
            \centering 
            \includegraphics[width=\textwidth]{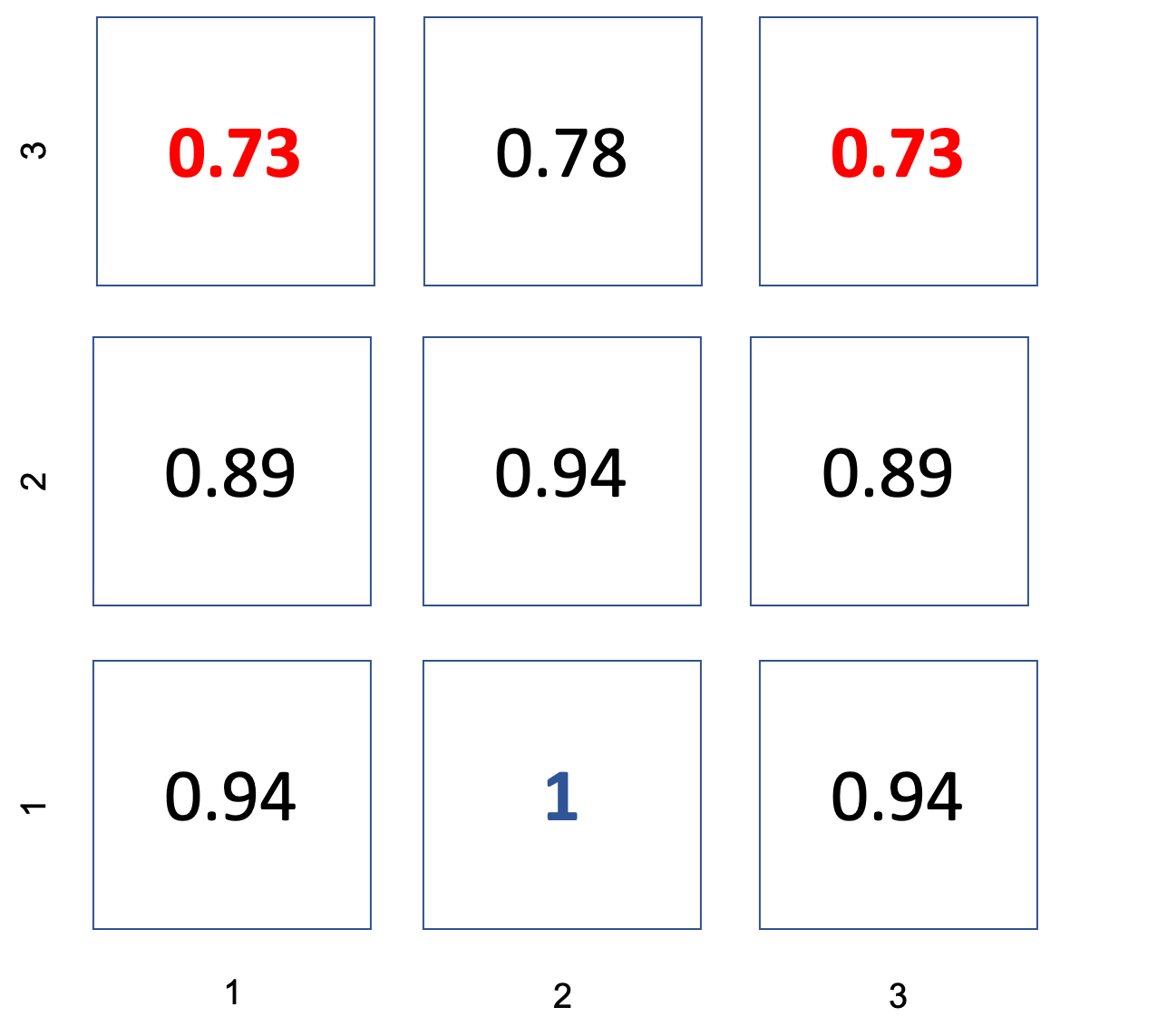}
            \caption[]%
            {{\small Similarities to $\mathbf{v}^P_{(2,2)}$, $\epsilon_{\mathcal{P}}>\epsilon_{\mathcal{D}}$.}}    
            \label{fig:fft3}
        \end{subfigure}

        \caption[  ]
        {\small Distribution of cosine similarities to hypervectors in $\mathcal{P}$ for every data hypervector after unbinding with hypervector $\mathbf{s}$.} 
        \label{fig:FFT}
    \end{figure}

In $\alpha_1$ and $\alpha_2$, the parameter of interest is $\epsilon_{\mathcal{P}}$, which is the bandwidth of HD-map $\mathcal{P}$ and is the hyperparameter of the algorithm. Consider the case where $\epsilon_{\mathcal{P}}>\epsilon_{\mathcal{D}}$. This means that the fidelity of the  similarity between hypervectors in HD-map  $\mathcal{P}$ is much coarser compared to the fidelity of the inter-hypervector similarity in the original vector space $\mathcal{D}$. In this case, all hypervectors from $\mathcal{D}$ after unbinding will be similar to the hypervector in $\mathcal{P}$, which was chosen for the update of seed hypervector $\mathbf{s}$ (e.g., $\mathbf{v}^P_{(p,q)}$ in Fig.
~\ref{fig:gridmapping}). Essentially, we will observe an effect of collapsing of all hypervectors in space $\mathcal{D}$ onto a single hypervector from space $\mathcal{P}$. This is demonstrated by a simulation in which hypervector  $\mathbf{s}$ was created as $\mathbf{s}=\mathbf{v}^D_{(1,2)} \circ \mathbf{v}^P_{(2,2)}$. The size of HD-map $\mathcal{P}$ is $5\times 5$.  Two simulations were performed: 1.) With the FPE bandwidth for encoding input data  being equal $0.2$, while the FPE bandwidth of HD-map was set to 0.03; and 2.) With the FPE bandwidth for encoding input data  being equal 0.2, while the FPE bandwidth of HD-map was set to 0.8.
Fig.~\ref{fig:fft1} and~\ref{fig:fft2} show the distribution of cosine similarities to all hypervectors of HD-map for every data hypervector after unbinding with hypervector $\mathbf{s}$ (\ref{eq:seedbind}) for the first and second simulation, respectively. Fig. \ref{fig:fft2} demonstrates the effect of collapsing of all points in $\mathcal{P}$ onto hypervector $\mathbf{v}^P_{(2,2)}$.  Fig. \ref{fig:fft3}  shows cosine similarities for every data hypervector after unbinding with hypervector $\mathbf{s}$ to the BMV in the second simulation (i.e., $\mathbf{v}^P_{(2,2)}$). 
Observe that the lowest similarities are for hypervectors $\mathbf{v}^D_{(1,3)}$ and $\mathbf{v}^D_{(3,3)}$, which are the farthest away from the hypervector $\mathbf{v}^D_{(1,2)}$ used to compute hypervector $\mathbf{s}$.  Therefore, following the FFTR heuristic, one of these hypervectors should be used in (\ref{eq:seedupdate}) to update hypervector $\mathbf{s}$, thus, creating a new point of attraction in $\mathcal{P}$. To summarize, the WMS procedure is as follows:

\begin{enumerate}

\item Initialize the lowest similarity variable $D_{min}=1$;
\item For each hypervector in $\mathcal{D}$ compute $p^*$ (\ref{eq:seedbind}) and search for the BMV in HD-map. Store the similarity to BMV $D_{BMV}$;
\item If $D_{BMV}<D_{min}$ then update the lowest similarity variable $D_{min}=D_{BMV}$. Store data hypervector as a candidate for the  update.  
\item Use the hypervector with lowest similarity for the update of hypervector $\mathbf{s}$ (\ref{eq:seedupdate}) on the current iteration.
 
\end{enumerate}


\subsection {Finding the target node on HD-map}
\label{sect:find}

The hypervector found during the WMS procedure must be anchored to a hypervector $\mathbf{p}_{\texttt{target}}$, encoding a node on HD-map. Since the WMS procedure finds a hypervector, which unbinds a hypervector  of HD-map with the lowest similarity, intuitively, it should be bound to a different hypervector. Using the FFTR heuristic, this new hypervector should be located further away from the current BMV in order to create a new point of attraction for the hypervector found by the WMS procedure and other hypervectors similar to it. 

\begin{figure}[t!]
        \centering
        \begin{subfigure}[b]{0.475\columnwidth}
            \centering
             \includegraphics[width=\textwidth]{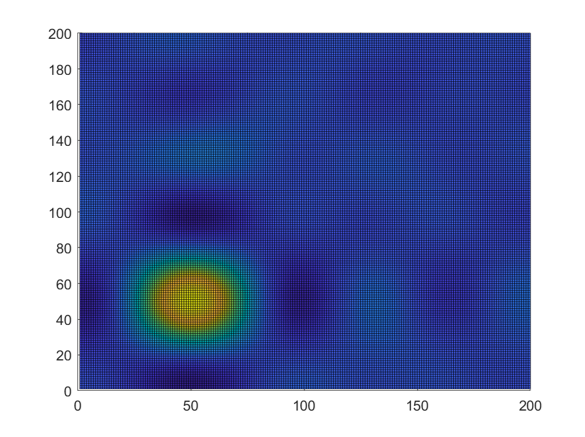}
            \caption[]%
            {{\small   $\epsilon_{\mathcal{P}}=0.03$.}}    
            \label{fig:fpeplace1}
        \end{subfigure}
        \hfill
        \begin{subfigure}[b]{0.475\columnwidth}  
            \centering 
            \includegraphics[width=\textwidth]{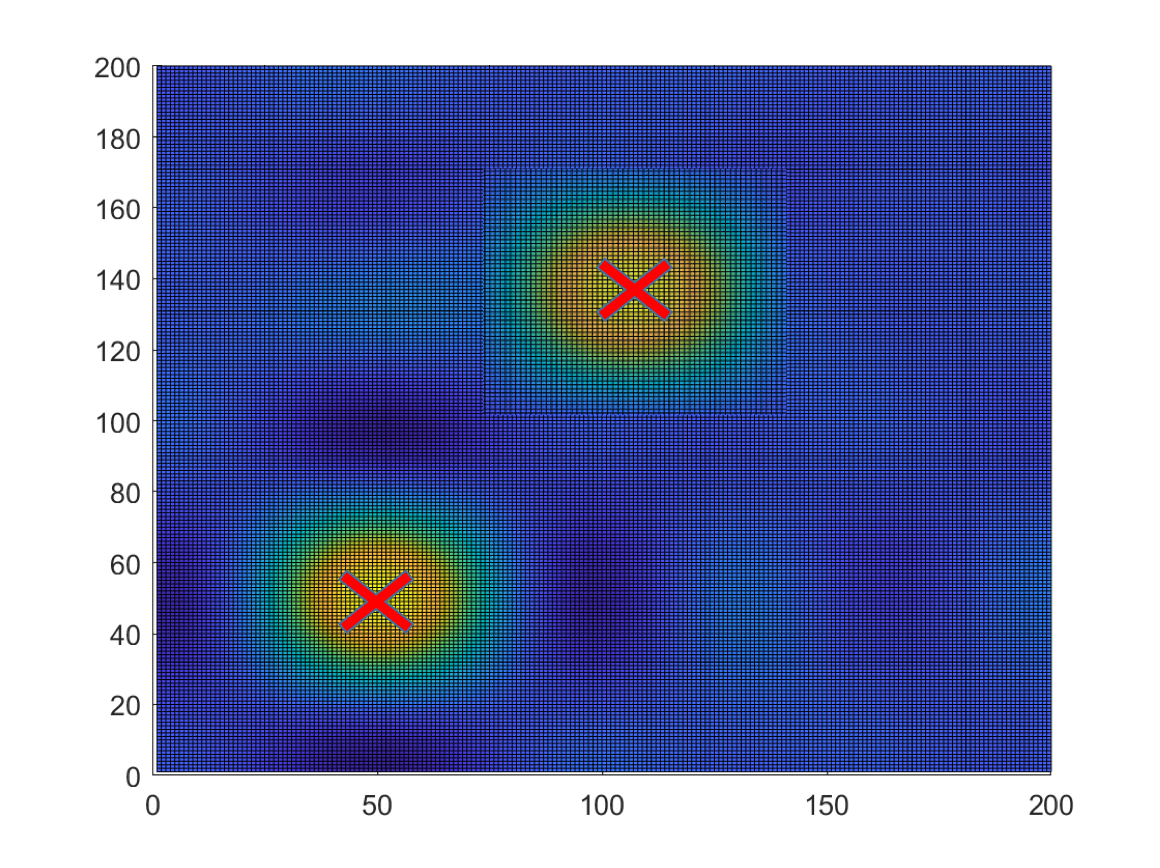}
            \caption[]%
            {{\small   $\epsilon_{\mathcal{P}}=0.03$.}}    
            \label{fig:fpeplace2}
        \end{subfigure}
        \vskip\baselineskip
        \begin{subfigure}[b]{0.475\columnwidth}   
            \centering 
            \includegraphics[width=\textwidth]{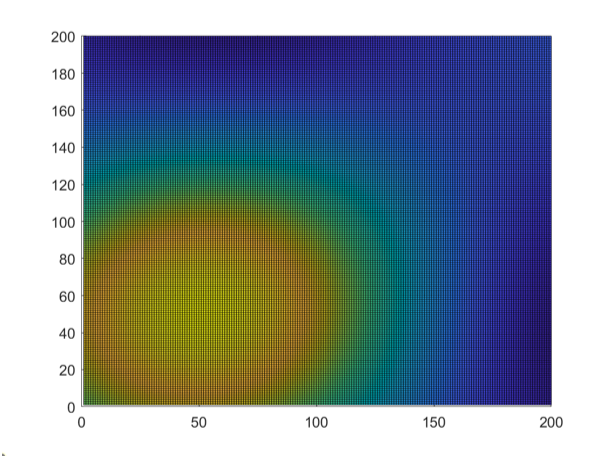}
            \caption[]%
            {{\small   $\epsilon_{\mathcal{P}}=0.008$.}}    
            \label{fig:fpeplace3}
        \end{subfigure}
        \hfill
        \begin{subfigure}[b]{0.475\columnwidth}   
            \centering 
            \includegraphics[width=\textwidth]{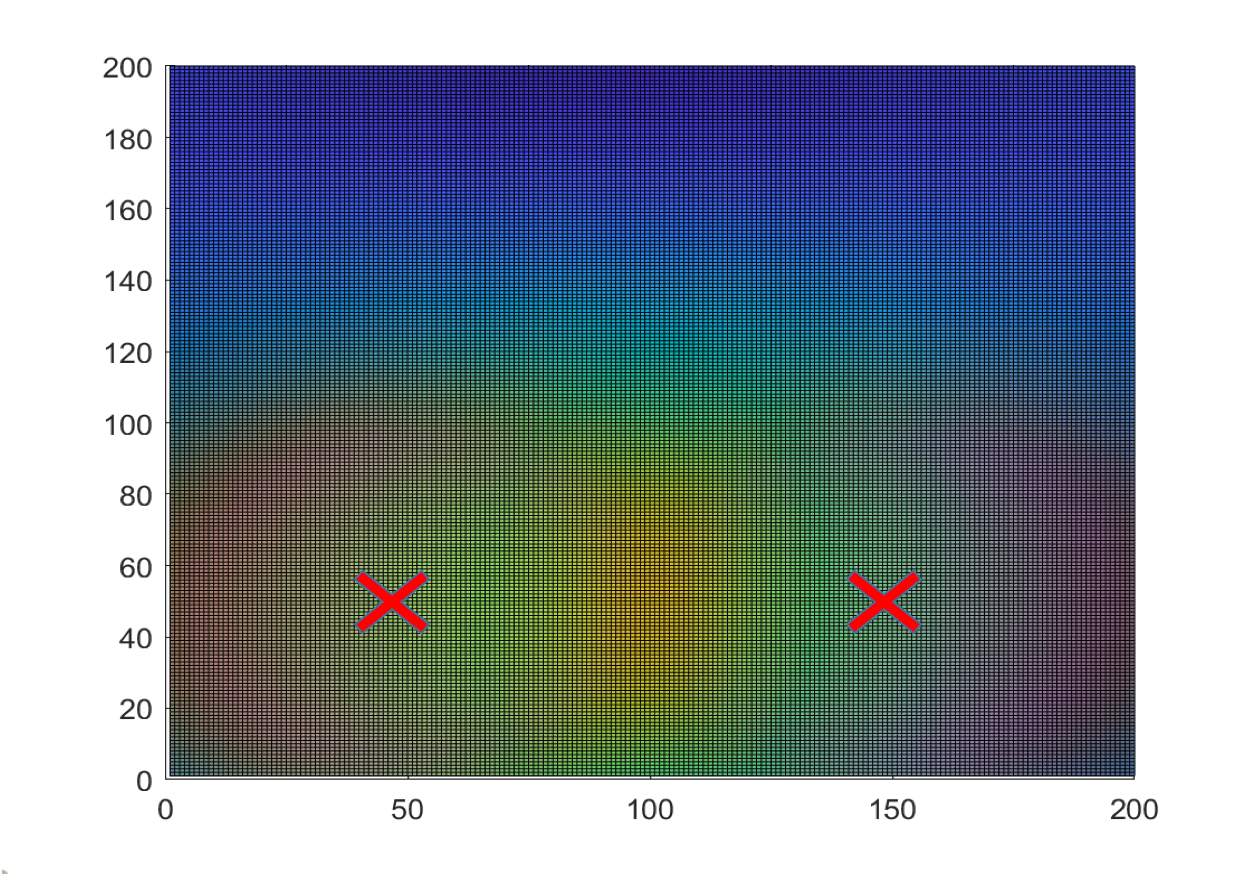}
            \caption[]%
            {{\small   $\epsilon_{\mathcal{P}}=0.008$.}}    
             \label{fig:fpeplace4}
        \end{subfigure}
        \caption[ ]
        {\small Distribution of cosine similarities on HD-map for different values of FPE bandwidth $\epsilon_{\mathcal{P}}$.} 
        \label{fig:fpeplace}
    \end{figure}

With the FPE encoding of HD-map's hypervectors the availability of farthest hypervectors to one selected hypervector is different depending on the choice of the bandwidth parameter for the same size of the map. For large values of $\epsilon_{\mathcal{P}}$ the similarity between hypervectors encoding neighboring nodes decays faster than for small values of $\epsilon_{\mathcal{P}}$, therefore, the number of dissimilar hypervectors is larger in the former case. This is demonstrated in Fig. \ref{fig:fpeplace1} and \ref{fig:fpeplace2} for $\epsilon_{\mathcal{P}}=0.03$ and Fig. \ref{fig:fpeplace3} and \ref{fig:fpeplace4} for $\epsilon_{\mathcal{P}}=0.008$ on $200\times200$ HD-map.

The simplest heuristic for finding the hypervector farthest to the given BMV  for HD-maps with large $\epsilon_{\mathcal{P}}$ is, therefore, a random selection of $\mathbf{p}_{\texttt{target}}$.
As we demonstrate in the next section, this heuristic leads to an adequate accuracy performance of Hyperseed in classification tasks. At the same time, obviously, the visualization of such projections is not very informative since it does not adequately display the internal disposition of classes. 

In the case of small values of $\epsilon_{\mathcal{P}}$, the number of farthest hypervectors on HD-map is limited and has to be selected according to a certain heuristic. When Hyperseed is used in visualization tasks, $\epsilon_{\mathcal{P}}$ is chosen such that most dissimilar hypervectors are located in the corners of HD-map. These corner nodes are then chosen as $\mathbf{p}_{\texttt{target}}$ during the update phase.

\begin{figure}[t!]
    \centering
    \includegraphics[width=6cm]{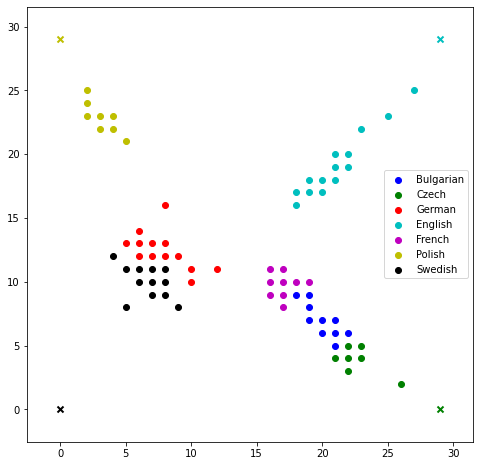}
    \caption{Projection of seven   European languages after four updates.  The chosen $\mathbf{p}_{\texttt{target}}$ hypervectors are four nodes (0,0), (20,0), (20,20) and (0, 20).  }
    \label{fig:lang_proj}
\end{figure}

Fig.~\ref{fig:lang_proj} demonstrates an instance of projections of a dataset containing collection of $n$-gram statistics from texts on seven European languages. The dataset is projected onto a $20\times20$ HD-map with $\epsilon_{\mathcal{P}}=0.008$. The complete experiment description follows in the next section.  In the figure, crosses in the corners of HD-map show the choice of target nodes during the update procedure. We observe  a semantically meaningful projections of languages.  The color of the crosses corresponds to the  class of the hypervector selected by the WMS procedure. One most important observation at this point is that the classes that were not used for the update of hypervector $\mathbf{s}$, e.g., Swedish, French, and Bulgarian languages,  emerged automatically and adequately projected.

 \subsection{The iterative Hyperseed algorithm}
 \label{sect:iterative}
 
\begin{figure}[t!]
    \centering
    \includegraphics[width=\columnwidth]{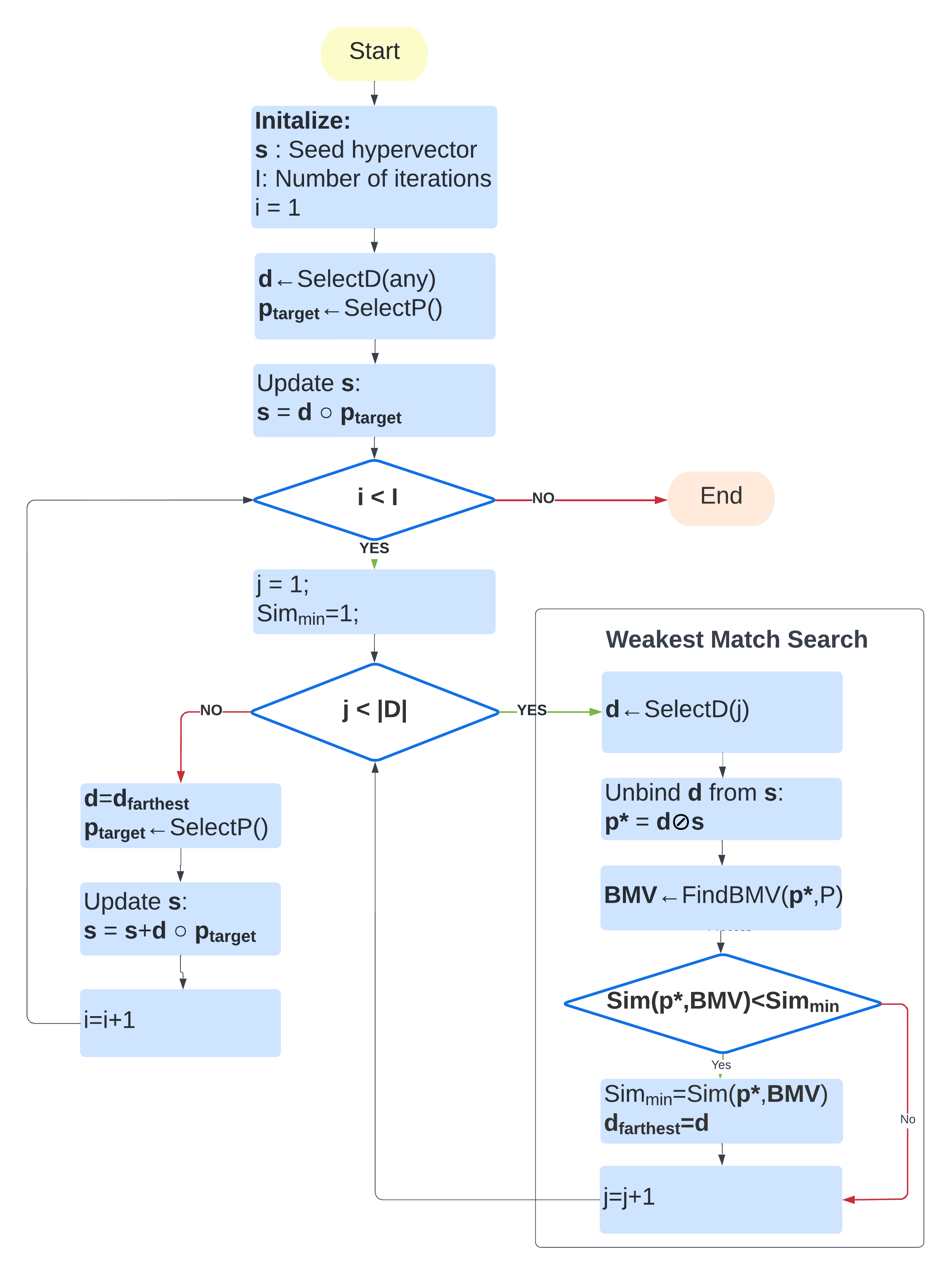}
    \caption{Flowchart of Hyperseed.  }
    \label{fig:flowchart}
\end{figure}

 Fig. \ref{fig:flowchart} displays all phases of the Hyperseed algorithm in a flowchart.  The hyperparameters of the algorithm are dimensionality of hypervectors $d$ and the number of iterations $I$ (i.e., the number of updates of hypervector $\mathbf{s}$). In the flowchart, function \texttt{SelectD()} returns an arbitrary hypervector from $\mathcal{D}$ if its argument is ``any''  or $j$-th hypervector from $\mathcal{D}$ when it is called with argument $j$. Function \texttt{SelectP()} returns a hypervector from $\mathcal{D}$ as described in the previous subsection. Function \texttt{FindBMV}($\mathbf{p}^*$,$\mathcal{P}$) performs a search in the associative memory storing hypervectors of $\mathcal{P}$ and returns the hypervector with the closest cosine similarity to $\mathbf{p}^*$. Function \texttt{Sim($\mathbf{p}^*$,$\mathbf{BMV}$)} computes the cosine similarity between the two hypervectors.
 
The computational complexity of Hyperseed is $\mathcal{O}(INd|\mathcal{P}|)$, where $I$ is the number of iterations (updates) of Hyperseed, $N$ is the number of training data hypervectors, $d$ is the dimensionality of hypervectors, and $|\mathcal{P}|$ is the size of HD-map. When implemented on neuromorphic hardware the search of the BMV happens in a constant time $\tau$, therefore, the time complexity of Hyperseed in this case is $O(IN)\cdot \tau$. The memory complexity, of course, depends on the size of HD-map which is $d \times |\mathcal{P}|$. 

The complexity of the SOM algorithm in the Winner-Takes-All  phase and in the weight matrix update procedures is  $\mathcal{O}(INd|SOM|)$.  Here, $I$ is the number of iterations of SOM algorithm, $N$ is the number of data samples , $|SOM|$ is the number of nodes in the SOM map (corresponds to $|\mathcal{P}|$ in Hyperseed) and $d$ is the number of neurons per node (corresponds to the dimensionality of hypervectors in Hyperseed). While the $\mathcal{O}$ complexities of the two algorithms are matching, Hyperseed uses a single vector operation in the update phase and substantially fewer number of iterations.

\section{Experiments and Results}
\label{sect:perf}

This section describes the results of the experimental evaluation of the proposed Hyperseed algorithm. 
Before elaborating on the details of the experimental evaluation, it is important to set  realistic expectations in order to correctly interpret the results. 
In particular, in the classification tasks, it would not be reasonable to expect a performance comparable to, e.g., deep learning-based models.
This is because classification is not the primary task for unsupervised learning approaches. As any other unsupervised learning algorithm, Hyperseed does not modify the input representations to make them more separable. We chose to evaluate the performance of Hyperseed on classification tasks mainly because visualization (as one of the main applications of the unsupervised learning) is subjective and it is hard to quantify its quality.

We report three illustrative cases: 1) unsupervised learning from one-shot demonstrations on six synthetic datasets; 2) classification of the Iris dataset; and 3) identification of $21$ languages using their $n$-gram statistics.  In all the experiments, we used dense complex FHRR representations \cite{plate1995holographic}  of varying dimensionality.  The Python implementation of the Hyperseed algorithm and the code necessary to reproduce all the experiments reported in this study are publicly available\footnote{Implementation of Hyperseed and the experiments, 2022. [Online.] Available: \url{https://github.com/eaoltu/hyperseed}}. 
\begin{table*}[tbh!]
    \caption{Comparison of projections and classification accuracy of Hyperseed vs. SOMs on FCPS datasets} \label{tab:hyperseed_fcps}
    \centering
    \begingroup
    \setlength{\tabcolsep}{6pt} 
    \renewcommand{\arraystretch}{2} 
    \begin{tabular}{ |c|c|c|c|c|c|c| } 
        \hline
        & Atom & Chain link & Engy time & Hepta & Two diamonds & Lsun 3D \\
        \cline{2-7}
        \raisebox{20pt}{\rotatebox{90}{Dataset}}
        &
        {\includegraphics[scale=0.25]{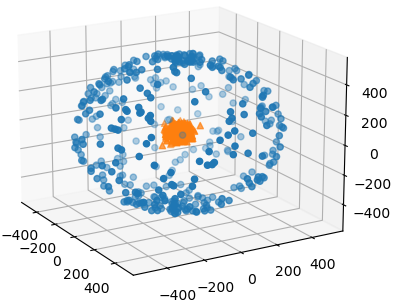}}
        &
        {\includegraphics[scale=0.26]{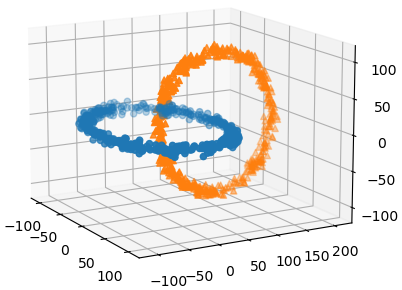}}
        &
        \raisebox{-3pt}{\includegraphics[scale=0.20]{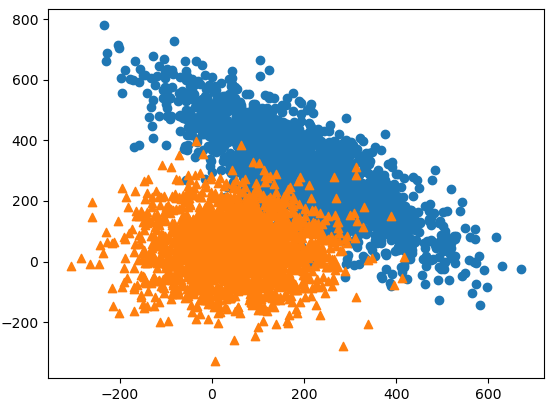}}
        &
        \raisebox{-3pt}{\includegraphics[scale=0.26]{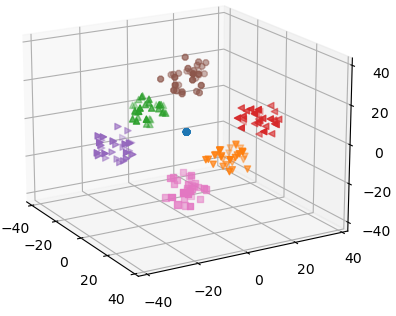}}
        &
        {\includegraphics[scale=0.19]{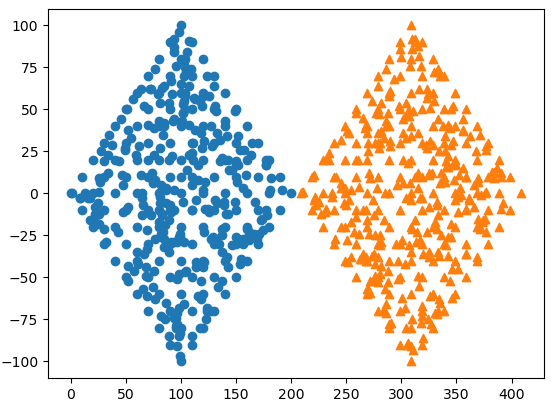}}
        &
        \raisebox{-3pt}{\includegraphics[scale=0.26]{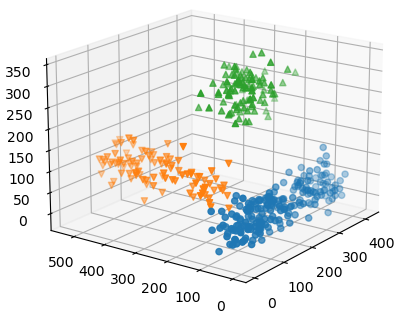}} \\
        \hline       
        \raisebox{10pt}{\rotatebox{90}{SOM}}
        &
        {\includegraphics[scale=0.185]{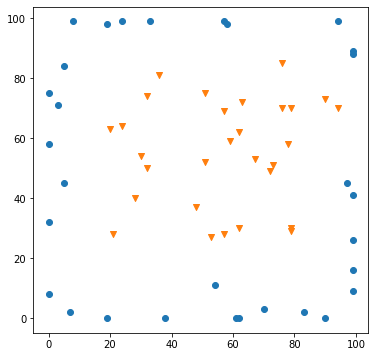}}
        &
        {\includegraphics[scale=0.185]{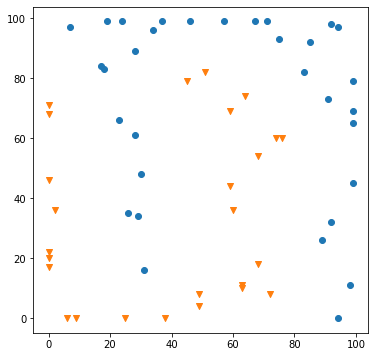}}
        &
        {\includegraphics[scale=0.185]{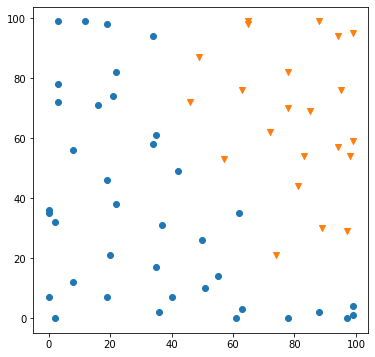}}
        &
        {\includegraphics[scale=0.185]{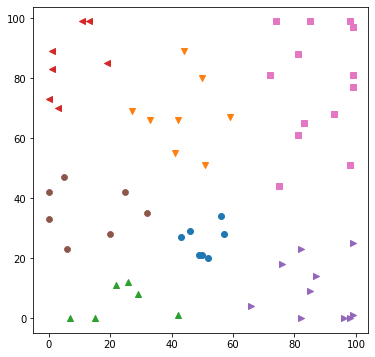}} 
        &
        {\includegraphics[scale=0.185]{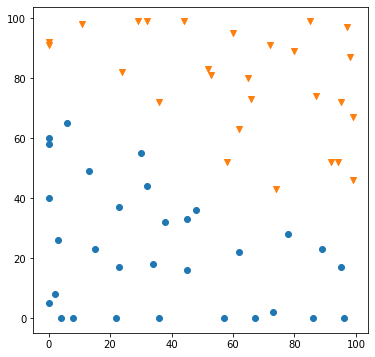}} 
        &
        {\includegraphics[scale=0.185]{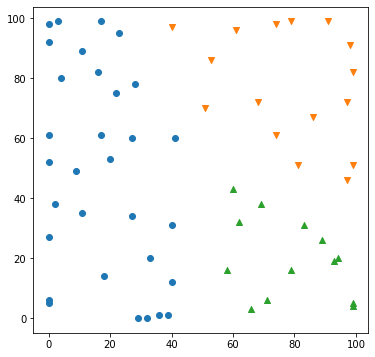}} \\ \cline{2-7}
        & Accuracy: 0.8878 & Accuracy: 0.9265 & Accuracy: 0.9528 & Accuracy: 0.9777 & Accuracy: 0.9774 & Accuracy: 0.9715 \\
        \hline
        \raisebox{5pt}{\rotatebox{90}{Hyperseed}}
        &
        {\includegraphics[scale=0.15]{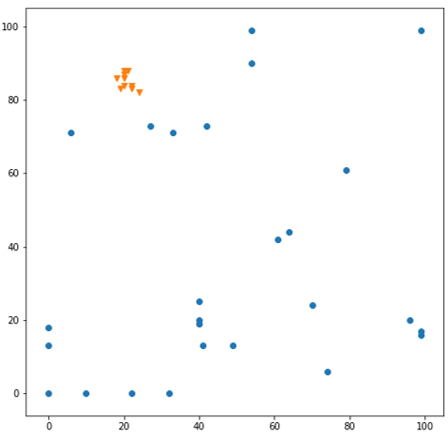}}
        &
        {\includegraphics[scale=0.15]{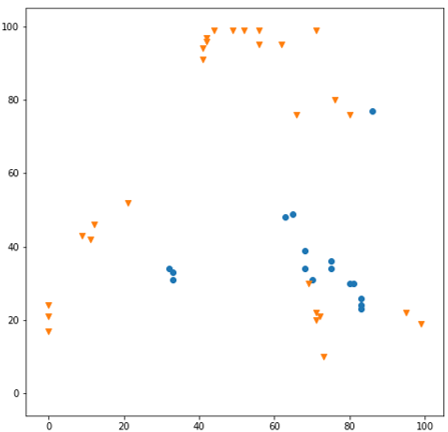}}
        &
        {\includegraphics[scale=0.15]{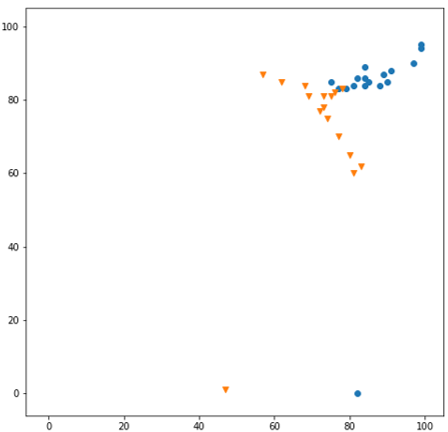}} 
        &
        {\includegraphics[scale=0.15]{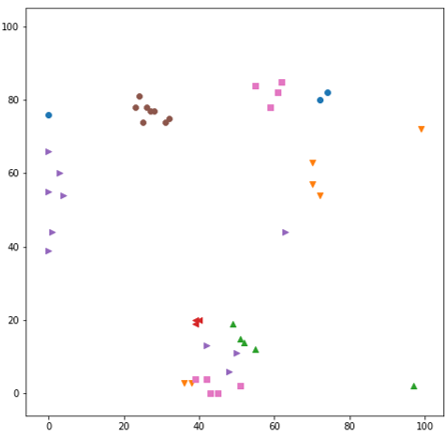}} 
        &
        {\includegraphics[scale=0.15]{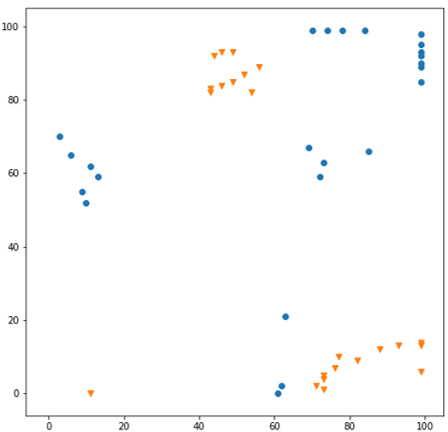}}
        &
        {\includegraphics[scale=0.15]{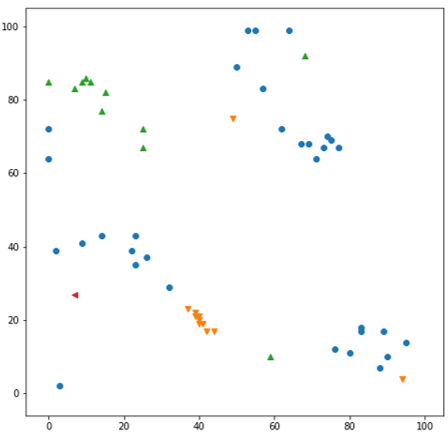}} \\ \cline{2-7}
        
        & Accuracy: 0.9821 & Accuracy: 0.9780 & Accuracy: 0.9071 & Accuracy: 0.9552 & Accuracy: 0.9902 & Accuracy: 0.9005 \\
        \hline
    \end{tabular}
    \endgroup
\end{table*}

\subsection{Data transformation to high-dimensional space}
\label{sect:encoding}
In the first two experiments, where input data are in the form of feature vectors of dimensionality $K$, the values of each feature $f_k, k \in [1,K]$ were normalized to range $[0,1]$. The interval $[0,1]$ was then split into $q$ quantization levels. 
For each feature a base hypervector of unit length $\mathbf{b}_k$ was randomly generated.
Then levels for the particular feature $\textbf{l}_i^k$ were encoded in FHRR representation using FPE~\cite{PlateNested1994, komer2020biologically, frady2021computing, FradyFunctionsNICE2022, Komer2019ANR}:  $\textbf{l}^k=\mathbf{b}_k^{\epsilon i}$, where $\epsilon \in \mathbb{R}$ represents the bandwidth parameter. The feature vector of  a data sample  was then represented as a single hypervector with the superposition operation: $\textbf{v}=\Sigma_{k \in K} \textbf{l}_i^k$.  

In the language identification experiments, $n$-gram statistics was represented as hypervectors following the procedure in \cite{RIJHK2015, KleykoBoostingSOM2019}. First, a bijection of the alphabet letters $i \in |\mathcal{A}|$ to random unitary atomic hypervectors $\mathbf{b}_i$ was created.  To encode the position of  character $i$ in the $n$-gram, the permutation operation was used on the corresponding atomic  hypervector $\mathbf{b}_i$. For example, the hypervector for character  ``b'' on a third position in a tri-gram is the corresponding  atomic hypervector rotated three times: $\rho^3(\mathbf{b}_{b})$.   An $n$-gram of size $n$ was encoded as binding of position based encoded hypervectors for corresponding characters. For example, a tri-gram  ``bdf''  was encoded as: $\rho(\mathbf{b}_{b}) \circ \rho^2( \mathbf{b}_{d})\circ \rho^3( \mathbf{b}_{f})$. 
The $n$-gram statistics for a given text sample was then encoded into a single hypervector through the superposition of hypervectors for all observed $n$-grams.

\subsection{Hyperseed for classification tasks}

The proposed Hyperseed algorithm is by definition an unsupervised learning algorithm, therefore, an extra mechanism is needed to use it in supervised tasks such as the considered Iris classification and language identification task. Once Hyperseed was trained, there is a need to assign labels to the best matching hypervectors in HD-map.  

Recall that Hyperseed is trained on a small subset of the available training data  as described in Section \ref{sect:iterative}. 
%
For the labeling process in the experiments presented below, the training data were presented to the trained Hyperseed HD-map for one full epoch that did not update the seed  hypervector $\mathbf{s}$. The labels of the training data were used to calculate  statistics for the BMV in HD-map. The nodes were assigned labels of the input samples that were prominent in the collected statistics. 

At the classification phase, hypervectors of the nodes with the assigned labels were stored in the memory. During the classification phase,  samples of the test data were used to assess the trained Hyperseed. For each sample in the test data, the BMV in HD-map was determined using the search procedure (Section~\ref{sect:search}). The test sample was then assigned the label of the closest labeled hypervector stored in the memory.

Accuracy was used as the main performance metric for evaluation and comparison of Hyperseed runs with different parameters. It should be re-emphasised that the focus of experiments was not on achieving the highest possible accuracy but on a comparative analysis of the Hyperseed algorithm.

\subsection{Experiment 1: The performance of Hyperseed on synthetic datasets with one-shot demonstration}

This experiment serves the purpose of highlighting the major property of the Hyperseed algorithm -- the capability of one-shot learning.  When talking about one-shot learning, one has to be careful with its definition. In many practical cases, a single example is obviously not enough for accurate inference, instead, it is reasonable to talk about learning from a limited number of data samples, i.e., few-shot learning. This is what we intend to gradually demonstrate with all the experiments in this section. 

In the first experiment, we fix the number of updates of seed hypervector $\mathbf{s}$ to one. This, essentially, boils down to randomly picking a sample from the particular training data, running the Hyperseed's update phase, and right after that performing labeling, classification on the corresponding test data, and the visualization. 

For this purpose, we used  synthetic datasets from  Fundamental Clustering Problems Suite (FCPS)~\cite{Ultsch05}. FCPS provides several non-linear but simple datasets that can be visualized in two or three dimensions, for elementary benchmarking of clustering and non-linear classification algorithms.

We selected six FCPS datasets that are most representative of the non-linearity, which we aim to learn using Hyperseed, they are: ``Atom'', ``Chain Link'', ``Engy Time'', ``Hepta'', ``Two Diamonds'', and ``Lsun 3D''. We repeated the experiment eight times. In each run, all hypervectors used for data encoding as well as for the operation of the Hyperseed algorithm were generated with a new seed used to initialize a pseudorandom generator. 

In Table~\ref{tab:hyperseed_fcps}, we present the results of this experiment in the form of the comparative evaluation of the projections by the conventional SOM and the projections produced by the Hyperseed algorithm, both visually and as the classification accuracy for each dataset. The sizes of the SOM grid and HD-map of the Hyperseed algorithm were the same $100\times 100$.
The first row in the table presents the visualization of the six selected datasets in their original two or three dimensional data space. The second row presents HD-map projections and classification accuracy of the conventional SOM, while the third row shows HD-map projections and classification accuracy of the Hyperseed algorithm.  

The primary observation in relation to the classification is that the Hyperseed algorithm provided average accuracy on a par  with the conventional SOM ($0.948$ versus $0.943$). 


The visualization of HD-map projections of Hyperseed are more representative of the topology preservation of the original data space, in comparison to the conventional SOM. In three out of six datasets, Engy Time, Hepta, and Lsun 3D, the topology preservation was directly comparable to the original dataset, where data samples of the same class were tightly clustered, in contrast to the conventional SOM, where these data samples were more scattered. For the remaining three datasets (Atom, Chain Link and Two Diamonds), although the topology preservation was not representative, the classification accuracy was high. This can be rationalized by the fixed 2D structure, which inhibits the complete visualization of the data space.

\subsection{Experiment 2: Iris classification with Hyperseed}

We continue the evaluation by exploring the details of Hyperseed's operations during several updates.  We used the Iris dataset with 150 samples of labeled data. The dataset  contains three classes of Iris flowers described by four real-valued features. Data were encoded into hypervectors as described in Section~\ref{sect:encoding}.  

To further highlight the capabilities of Hypreseed to learn from few-shots, we decided to split the Iris dataset such that the size of the test data was larger than the size of the training data. The results for Hyperseed in this section were obtained for the $20\% / 80\%$ split for training and test data, respectively. In order to provide a benchmark performance, we trained the conventional SOM with a $30 \times 30$ grid on different splits of the Iris dataset and counted the number of iterations it required the SOM to reach $95\%$ accuracy. The experiments with the conventional SOM were repeated 10 times and the maximum accuracy across runs was recorded. Table~\ref{tab:num_iter} reports the results. 
One could see that for the 20/80 split the conventional SOM failed to achieve the target accuracy. Increasing the size of the training data allowed SOM reaching the target accuracy. The number of required iterations, however, was significant (the lowest was 200 for the 80/20 split).
The maximum number of iterations of Hyperseed was set to 3 and 6. 
This means that algorithm performed 90  (3 times 30 samples of the training data) and 180 (6 times 30 samples of the training data) searches for the BMV and only 3 (and correspondingly 6) updates of seed hypervector $\mathbf{s}$. This has to be compared to $200 \times 120$ of searches and updates of the conventional SOM in the 80/20 split case. Each experiment (with three and six updates) was run ten times with different seeds. 

The target nodes at each update were pre-selected to be (15,15) - for the first update, (20,20) - for the second update, (10,10) - for the third update, (5,5) - for the fourth update, (25,25) - for the fifth update, and (5,25) - for the sixth update.  These nodes are marked by red crosses in the visualizations.  This highlights again the importance of the target node selection rule for visually adequate projections. In this experiment, however, the adequate visualization was not important since our focus was on the performance characteristics of the Hyperseed algorithm in the classification task. 
Table~\ref{tab:hyperseed_comp1} shows the results of the experiment. For the fair comparison with the conventional SOM here we also select the best performance across runs.     

\begin{table}[t!]
    \caption{Number of iterations of the conventional SOM required to reach $95\%$ accuracy for different splits of Iris.} \label{tab:num_iter}
    \centering
    \renewcommand{\arraystretch}{2} 
    \begin{tabular}{ |p{4cm}|c|c|c|c| } 
        \hline
        Data set split (training/testing), \% & 20/80 & 40/60 & 60/40 & 80/20 \\ \hline
        Number of updates  
        & - & 2500 & 600 & 200 \\
                \hline

    \end{tabular}
\end{table}

\begin{table}[t]
    \caption{Comparison of the Hyperseed projections with different random seeds for the Iris dataset.} \label{tab:hyperseed_comp1}
    \centering
    \begingroup
    \setlength{\tabcolsep}{6pt} 
    \renewcommand{\arraystretch}{2} 
    \begin{tabular}{ |c|c|c| } 
        \hline
        
        &
        \# updates: 3, Acc: 0.92 & \# updates: 6, Acc: 0.95  \\

        \hline
        {\rotatebox[origin=c]{90}{Train Projection}}
        &
        \raisebox{-45pt}{\includegraphics[scale=0.2]{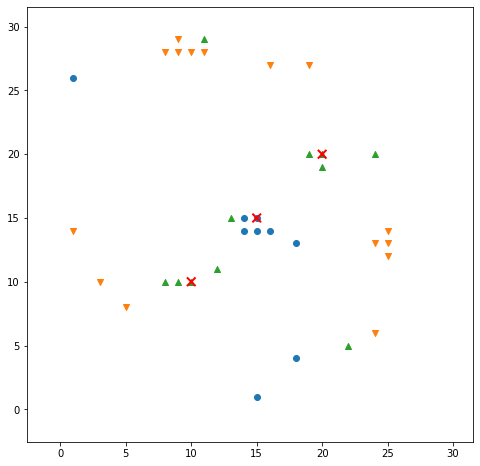}}
        &
        \raisebox{-45pt}{\includegraphics[scale=0.2]{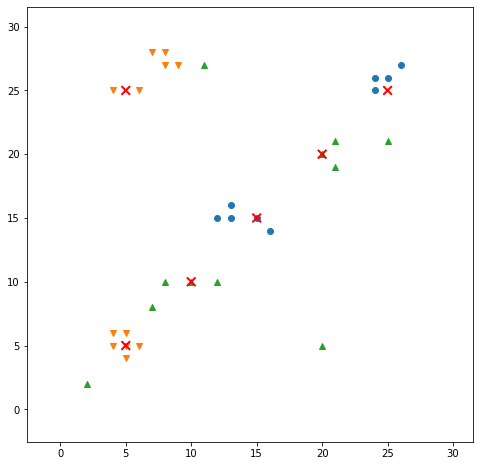}}
         \\
    
        \hline
        \rotatebox[origin=c]{90}{Test Projection}
        &
        \raisebox{-45pt}{\includegraphics[scale=0.2]{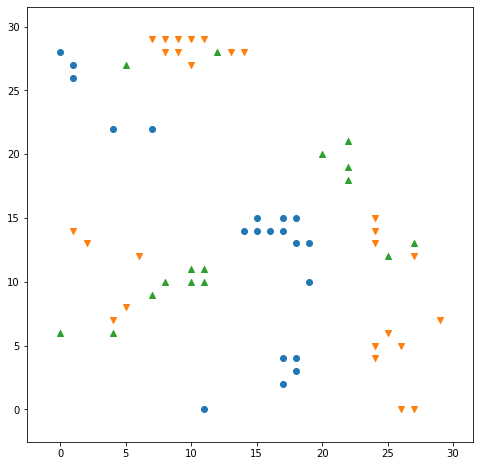}} 
        &
        \raisebox{-45pt}{\includegraphics[scale=0.2]{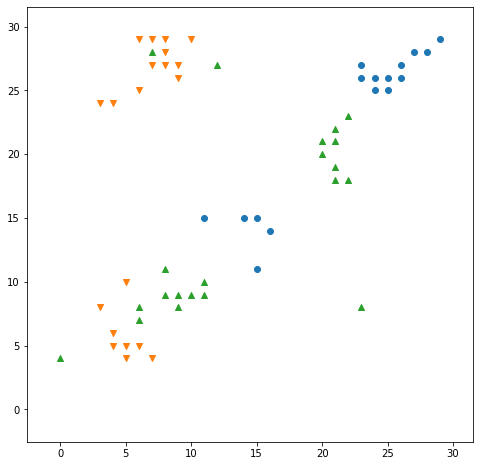}}\\
        
        \hline
    \end{tabular}
    \endgroup
\end{table} 

The first interesting observation comes in the case of three updates. 
We selected the best runs, which resulted in the highest accuracy on the test data ($0.93$). The projection of the training data show that out of three updates, in total one update was done for the first class (blue circles) and two updates were done for the second class. 
Thus, the cluster for the third class emerged automatically. 

\begin{figure}[t!]
    \centering
    \includegraphics[width=8cm]{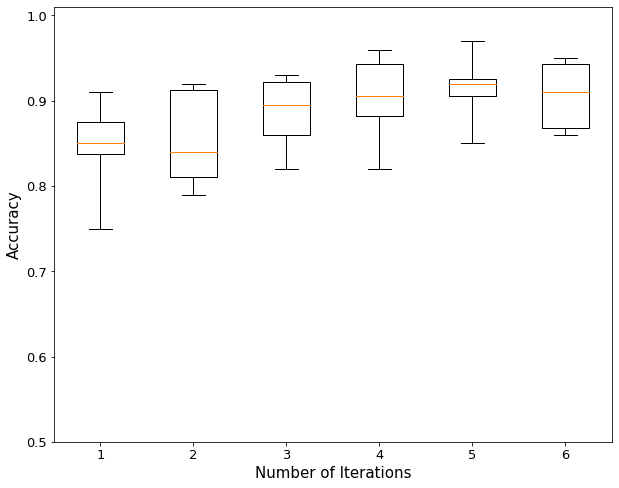}
    \caption{Number of iterations against the accuracy ($d=500$) of Hyperseed on the Iris dataset. }
    \label{fig:accVsIterIRIS}
\end{figure}

\begin{figure}[t!]
    \centering
    \includegraphics[width=8cm]{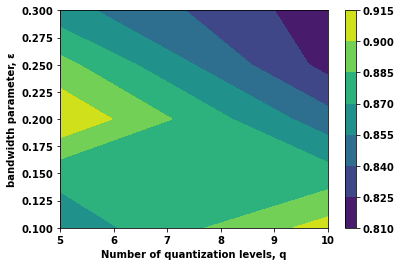}
    \caption{Classification accuracy against the choice of FPE hyperparameters for data encoding: bandwidth and the number of quantization levels. Dimensionality of hypervectors was set to $d=500$.}
    \label{fig:accVSqVSbw}
\end{figure}

Another important observation comes from the relative placement of the data samples (both from the training and test data) on HD-map. For the Iris dataset, it is known that the second and third classes are very similar to each other, which manifests in misclassification of some of their samples. Here, we see that this was, indeed, the case (orange triangles are very close to green triangles in several nodes of HD-map). 
However, in the case of Hyperseed this proximity did not lead to large degradation of the classification accuracy. This is because each point in HD-map attracted similar samples.
Next, Fig.~\ref{fig:accVsIterIRIS} shows the accuracy of Hyperseed when increasing the number of updates. On average, the classification accuracy increased with more updates of seed hypervector $\mathbf{s}$.
Also, in Fig.~\ref{fig:accVSqVSbw} we demonstrate the tradeoff between the choice of hyperparameters for the FPE encoding of input data (bandwidth $\epsilon$ and the number of quantization levels, $q$)  and the classification accuracy.  
We observed that the best accuracy was achieved when $\epsilon$ and $q$ were in the inverse relationship, that is when $\epsilon q = 1$.

    \begin{figure*}[t!]
        \centering
        \begin{subfigure}[b]{0.475\textwidth}
            \centering
             \includegraphics[width=\textwidth]{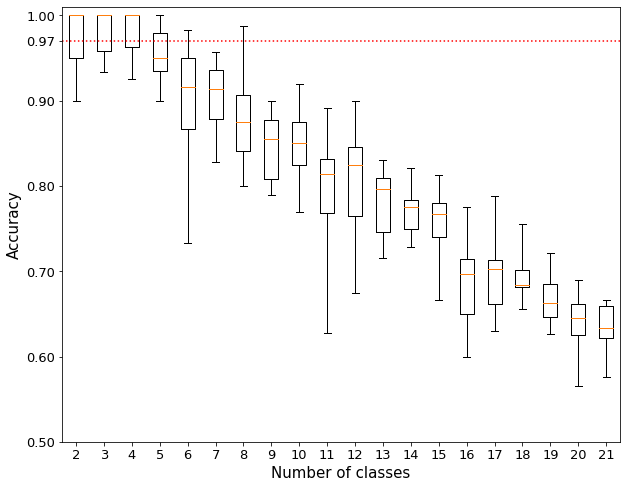}
            \caption[]%
            {{\small Accuracy with one seed hypervector $\mathbf{s}$  using the original update rule from Section~\ref{sect:update}, $d=5000$.}}    
            \label{fig:single}
        \end{subfigure}
        \hfill
        \begin{subfigure}[b]{0.475\textwidth}  
            \centering 
            \includegraphics[width=\textwidth]{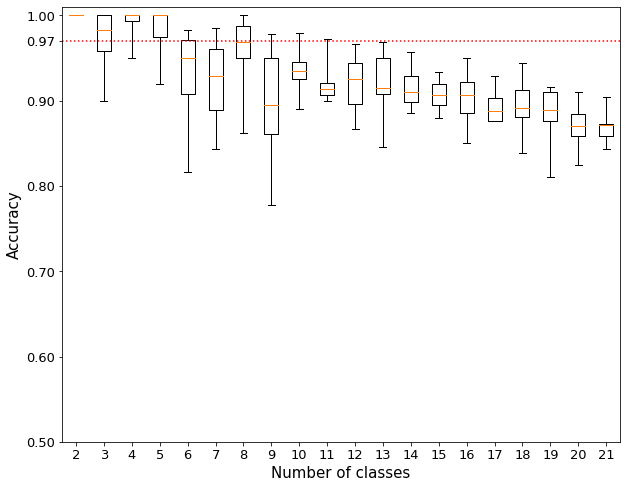}
            \caption[]%
            {{\small Number of classes vs. accuracy with original and modified learning phases ($d=5000$).}}    
            \label{fig:ten}
        \end{subfigure}
        \vskip\baselineskip
        \begin{subfigure}[b]{0.475\textwidth}   
            \centering 
            \includegraphics[width=\textwidth]{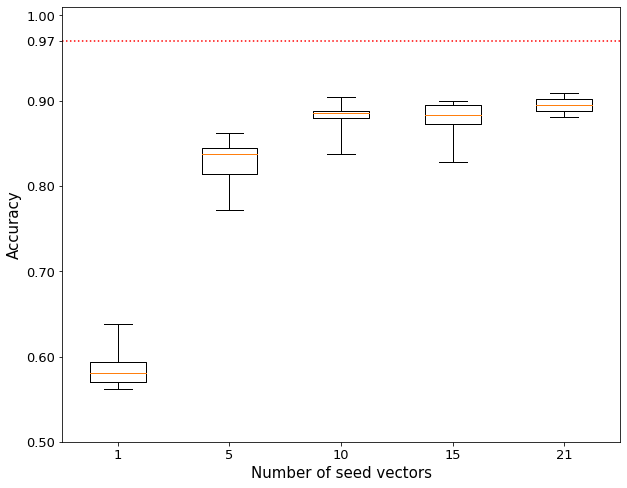}
            \caption[]%
            {{\small Number of $\mathbf{s}$ hypervectors vs. accuracy ($d=5000$).}}    
            \label{fig:num_seed}
        \end{subfigure}
        \hfill
        \begin{subfigure}[b]{0.475\textwidth}   
            \centering 
            \includegraphics[width=\textwidth]{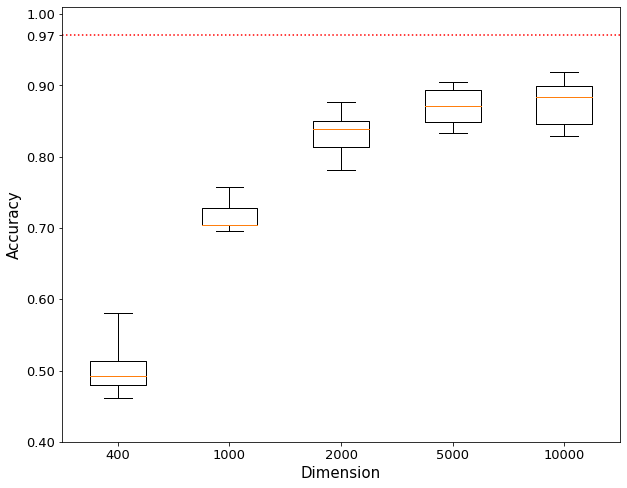}
            \caption[]%
            {{\small Dimensionality of hypervectors vs. accuracy (Number of $\mathbf{s}$ hypervectors is 10).}}    
             \label{fig:dimensionality}
        \end{subfigure}
        \caption[ TResults of Experiment 3. Performance of Hyperseed in 21 languages classification task. ]
        {\small The performance of Hyperseed in the task of language identification.} 
        \label{fig:Experiment3}
    \end{figure*}

\subsection{Experiment 3: Training Hyperseed on $n$-grams in language identification task}
While the previous two  experiments were used to get first  insights on  the major properties of Hyperseed, with this experiment we intend to demonstrate its performance on a larger scale problem. 
We will use it for the classification task of 21 European languages using $n$-gram statistics. The list of languages is as follows: Bulgarian, Czech, Danish, German, Greek, English,
Estonian, Finnish, French, Hungarian, Italian, Latvian, Lithuanian, Dutch, Polish, Portuguese, Romanian, Slovak, Slovene, Spanish, Swedish. 
The training data is based on the Wortschatz Corpora \cite{LANGDATA}. 
The average size of each language corpus in the training data was $1085637.3 \pm 121904.1$ symbols. 
In this study, we use the method for encoding $n$-gram statistics into hypervectors from~\cite{KleykoBoostingSOM2019}, where it was used as an input to the conventional SOM. Each original language corpus was divided into samples where the length of each sample was set to $1000$ symbols. 
The total number of samples in the training data was $22791$.

The test data also contains samples from the same languages and is based on the Europarl Parallel Corpus\footnote{Available online at \url{http://www.statmt.org/europarl/}. 
}.
The total number of samples in the test data was $21000$, where each language was represented with $1000$ samples. 
Each sample in the test data corresponded to a single sentence. 
The average size of a sample in the test data was $150.3 \pm 89.5$ symbols.

The data for each language was pre-processed such that the text included only lower case letters and spaces. 
All punctuation was removed. 
Lastly, all text used the 26-letter ISO basic Latin alphabet, i.e., the alphabet for both training and test data was the same and it included $27$ symbols. 
For each text sample, the $n$-gram statistics transformed to hypervectors was obtained, which was then used as input  when training or testing Hyperseed.
Since each sample was pre-processed to use the alphabet of only $a=27$ symbols, the conventional $n$-gram statistics input was $27^{n}$ dimensional. In the experiment, we used tri-grams, therefore, the conventional representation was $19,683$ dimensional. The dimensionality of the mapped  $n$-gram statistics into hypervectors as described in Section~\ref{sect:encoding} depends on the dimensionality of the hypervectors $d$. The results reported below  were obtained for different dimensionalities in the range [400, 10000], which corresponds to the dimensionality reduction of the original representation space from $49$ fold (for $d=400$) to $2$ fold (for $d=10,000$). In this experiment, we used $100\times100$ HD-map.


The first investigation was performed with dimensionality of hypervectors $d=5000$ ($4$ fold dimensionality reduction). 
In this experiment, we exposed Hyperseed to a different number of classes from the original dataset. The number of Hyperseed updates in each case was set relative to the number of classes at hand. Importantly, this heuristic for choosing the number of updates was adopted for automating experiments only. It only reflects the desire of keeping the number of iterations low.  In the unsupervised learning context, the information about the number of classes is, obviously, unavailable. 

The experiment was repeated eight times for each number of classes, each time selecting a different subset of languages for Hyperseed training. Fig.~\ref{fig:single} shows the results. The reference accuracy 0.97 (the red line) is the accuracy obtained with the conventional tri-gram statistics representation reported in \cite{HDenergy, RIJHK2015}  using the nearest neighbor classifier.  
The main observation to make from this investigation is that the performance of Hyperseed dropped as it was exposed to larger number of classes. The explanation to this is connected to the finite capacity of hypervectors in terms of the number of hypervectors one can superimpose together while keeping the sufficient accuracy of retrieving them back~\cite{FradyCapacity2018,KleykoPerceptron2020}. As we discussed above, each update to seed hypervector $\mathbf{s}$ introduces noise to the previous updates. That is why the number of updates in Hyperseed needs to be kept low.  

\subsubsection{Modified Hyperseed learning phase}
To mitigate the problem of the reduced accuracy on large problems (in terms of the number of distinct classes) the learning phase of Hyperseed was modified. 

Instead of having a single seed hypervector, it is proposed to use $N$ vectors $\mathbf{s}_i, i=1..N$, where $N$ becomes another hyperparameter of Hyperseed. During the update procedure, these hypervectors will be updated in a round-robin manner in order to keep the balanced number of updates per a seed hypervector.

During the search of the BMV in either the WMS or test procedures, the binding in (\ref{eq:bind}) is now computed for the current  input hypervector and all seed hypervectors  $\mathbf{s}_i$'s.  
Hypervector $\mathbf{p}$ with the highest cosine similarity across all results of unbinding with all $\mathbf{s}_i$ is selected as the BMV.

The number of iterations (and as the result the number of updates) with the new update phase should be scaled such that the number of updates per $\mathbf{s}_i$ is approximately the same. For example, in the experiments with  $N=10$ the number of iterations was configured to $30$ to allow for three updates per $\mathbf{s}_i$. 
Fig.~\ref{fig:ten} demonstrates significant performance improvement with the modified learning phase. The maximum performance of Hyperseed in $21$ classes case is $0.91$ after $30$ iterations.

Next, we are interested in the effect of the number of seed hypervectors on the classification performance of Hyperseed. 
Fig.~\ref{fig:num_seed} shows the classification accuracy in the case of 21 classes for different number of seed hypervectors used in the modified learning phase. The main observation here was that the performance stabilized after a certain value of this parameter. 
In our case, there was no significant increase in the accuracy after using ten seed hypervectors.

\subsubsection{Hyperseed performance for different dimensionalities of hypervectors}

Finally, we  are interested in the effect of the dimensionality of hypervectors on the performance of Hyperseed. We performed an experiment with ten seed hypervectors and varied the dimensionality of the hypervectors used by the algorithm from 400  until 10000. The results are depicted in Fig.~\ref{fig:dimensionality}. The main observation to make was that the classification accuracy on small dimensionalities was low, as expected. On the positive side, it was substantially higher than the random choice, which is $0.04$ in this case. The reason for this is rather clear and it is connected to the dimensionality of the non-distributed representations, which is high ($19,683$). With only $400$ dimensions and the adopted encoding procedure of the input data, we operated well above the capacity of the hypervectors, this lead to high inter-class similarity. 
Increasing the dimensionality allowed addressing this issue and resulted in better accuracy. It turned out that in the case of the language identification $5000$ was the optimal dimensionality for obtaining high-quality performance. It is worth noting that the accuracy of Hyperseed was still  lower by $0.07$ compared to the baseline. It was, however, not expected that it would necessarily achieve higher accuracy compared to the supervised methods. 

\section{Discussion}
\label{sect:discussion}

Having presented the Hyperseed algorithm and its empirical evaluation, it is now pertinent to discuss the following aspects that require further investigation, Hyperseed on neuromorphic hardware, performance comparison of embeddings and limitations of Hyperseed. 

\subsection{Hyperseed on neuromorphic hardware}

The Hyperseed algorithm from the start was designed targeting an implementation on the neuromorphic hardware. This target departs from  recent developments within the Intel neuromorphic research community, where VSA is promoted as an algebraic framework for the development of algorithms on Intel's Loihi \cite{Loihi18, TPAM,  Frady20_KNN}. 

Since the main focus of this article is on the algorithmic aspects of Hyperseed, we resort to making rather high level links to a neuromorphic realization of algorithm's operations and present an evaluation of its computational bottleneck using the existing neuromorphic implementation.

Both HRR and FHRR representations could be mapped onto activities of spiking neurons. 
In the case of FHRR, the phase of components is used for phase-to-spike-timing mapping.
In this way, FHRR representation does not result in higher memory footprint as in the case of the CPU implementation. 
VSA operations are realized in either Resonate-and-Fire neurons~\cite{TPAM} or in Leaky Integrate and Fire neurons~\cite{RennerBinding2022, RennerVisualScene}.

In the case of HRR, real-valued components are used in spike-time latency code, where
earlier spikes represent larger magnitudes. 
VSA operations can realized by Leaky Integrate and Fire neurons. 
In this article, we use the realization of the dot product calculation  presented in \cite{ Frady20_KNN} (also based on Leaky Integrate and Fire neurons) in order to demonstrate the feasibility of neuromorphic implementation of computational bottleneck of Hyperseed -- the search for the BMV in HD-map for an unbound noisy hypervector $\textbf{p}^*$ resulting from~(\ref{eq:seedbind}).

To  measure the performance of the Hyperseed's search procedure in the language identification task  HD-maps of various sizes (starting from $30\times30$ and incrementing the grid size by $10$ along each axis until $90\times90$) were generated as in Section~\ref{sect:hdMap}.  The hypervectors of each HD-map were then used for mapping their values onto spiking activity of the $k$NN reference base  on  Intel's Loihi-based Nahuku-32 neuromorphic system.  This operation was performed only once as part of the initialization since HD-map remains unchanged for the life-time of Hyperseed. Therefore, the time to construct the $k$NN reference base was not taken into account in the run-time performance evaluation.

For the experiment we chose the case of identifying five randomly selected languages with a single  seed hypervector $\mathbf{s}$ as the reference scenario. The original dimensionality of the hypervectors used for the encoding of the input data as well as for all other hypervectors of the Hyperseed algorithm was $d=10000$.  
The reference scenario accuracy of Hyperseed obtained on a CPU was $0.84$. 

$k$NN on Loihi was used to model the search operation during the labeling and testing process.  To do this, seed hypervector $\mathbf{s}$ was pre-trained offline on the CPU as described in Section~\ref{sect:update}. For the labeling process, the binding  of all training and test data with the trained $\mathbf{s}$ was performed and the results (the noisy versions of the BMVs) were used as queries to the $k$NN reference base storing HD-map on Loihi.
The dimensionality of the existing Loihi implementation of $k$NN is  $d_{kNN}<512$, which is a platform specific limit. 
Note, that in VSA the dimensionality of hypervectors is connected to the information capacity of the superposition. In Hyperseed, every update of seed hypervector $\mathbf{s}$ increases the cross-talk noise to previous bindings $\mathbf{s}\circ \mathbf{p}_{\texttt{target}}$. Fig.~\ref{fig:single} demonstrated the accuracy degradation for dimensionality $d=5000$ with the increase of the number of updates of $\mathbf{s}$. For smaller dimensionalities, the number of updates for maintaining the acceptable accuracy is even lower. Therefore, in this experiment the training of Hyperseed was done on higher dimensionalities and then the dimensionality was reduced using principal component analysis to $d=400$ to meet the current limitations of the existing implementation. 

To label BMVs, the training data (after binding with $\mathbf{s}$) were used as queries to the $k$NN reference base and the top-1 index for each query (the earliest fired output neuron) was recorded. After that, the labeling was performed on the CPU using the procedure described above. To compute the accuracy, the test data hypervectors (also after binding with $\mathbf{s}$) were used as queries to the $k$NN reference base. The top-1 index was recorded and used to compute the accuracy against the list of the labeled indices.   

The average measured accuracy on Loihi was $0.84$, which matched the reference accuracy on the CPU implementation of Hyperseed. This means that the neuromorphic implementation of HD-map and the calculation of similarity did not introduce sensible errors.

\begin{figure}[t!]
    \centering
    \includegraphics[width=8cm]{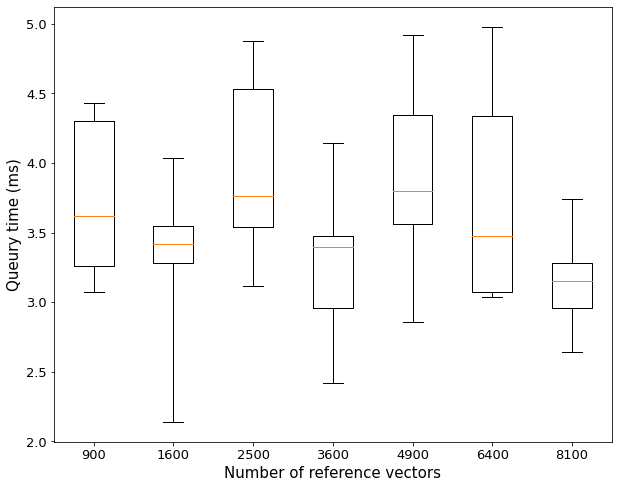}
    \caption{Time of querying a noisy BMV in HD-map implemented as the $k$NN reference base  against the size of HD-map in the number of reference hypervectors.}
    \label{fig:qTimeVsNumberOfRefVecs}
\end{figure}

Next, in addition to the accuracy we also measured the query time to the $k$NN reference base for different sizes of HD-map. The main outcome of this experiment is illustrated by Fig. \ref{fig:qTimeVsNumberOfRefVecs}. It shows the  computational benefit  of implementing the bottleneck operation of Hyperseed on in the neuromorphic hardware: due to parallel, power-efficient computation of the dot product in Loihi, Hyperseed, as expected, was empowered with  constant-time search for different sizes of HD-map.

  

\subsection{Performance comparison of embeddings and distributed representations}
 While a large body of knowledge on methods for encoding data into hypervectors is accumulated throughout the years, the problem is still considered as of the primary importance in the area of VSA. Hyperseed in this respect offers a playground for comparing different embeddings as the embeddings of high quality  lead to higher accuracy in classification tasks. It is particularly important to develop the operations of Hyperseed on sparse representations. 
 
 \subsection{Limitations of Hyperseed and future developments}
Hyperseed when using 2D HD-map as evaluated in this article is limited in its capability to describe complex manifold structures. In this sense the algorithm does not show advantages over Self-Organizing Maps, which have similar limitations. This case was chosen to demonstrate the feasibility of implementing non-trivial learning functionality with straightforward VSA operations, which in itself is an original research contribution. However, Hyperseed is in fact scalable in terms of two other aspects:  1) topologies of higher dimensionalities than two as it does not require updates of the hypervectors in HD-map and 2) topologies of other structures than regular grid due to the generality of FPE encoding. The investigation of this capability is part of an ongoing work on Hyperseed extension.

\section{Conclusions}
\label{sect:conclusions}

The increasing accumulation of unstructured and unlabelled big data initiated a renewed interest in unsupervised machine learning algorithms that are able to capitalize on computational efficiencies of biologically-inspired neuromorphic hardware. 
In this article, we presented the Hyperseed algorithm that addresses these challenges through the manifestation of a novel unsupervised  learning approach that leverages Vector Symbolic Architectures for fast learning from only few input vectors and single vector operation learning rule implementation. A further novelty is that it implements the entire learning pipeline purely in terms of operations of Vector Symbolic Architectures. Hyperseed has been empirically evaluated across diverse scenarios: synthetic datasets from Fundamental Clustering Problems Suite, benchmark classification using the Iris dataset, and the more practical classification of $21$ European languages using their $n$-gram statistics. 
As future work, we will work on the adaptation of the Hyperseed algorithm for neuromorphic hardware.

\bibliographystyle{IEEEtran} 

\bibliography{bica}


\begin{IEEEbiography}[{\includegraphics[width=1in,height=1.25in,clip,keepaspectratio]{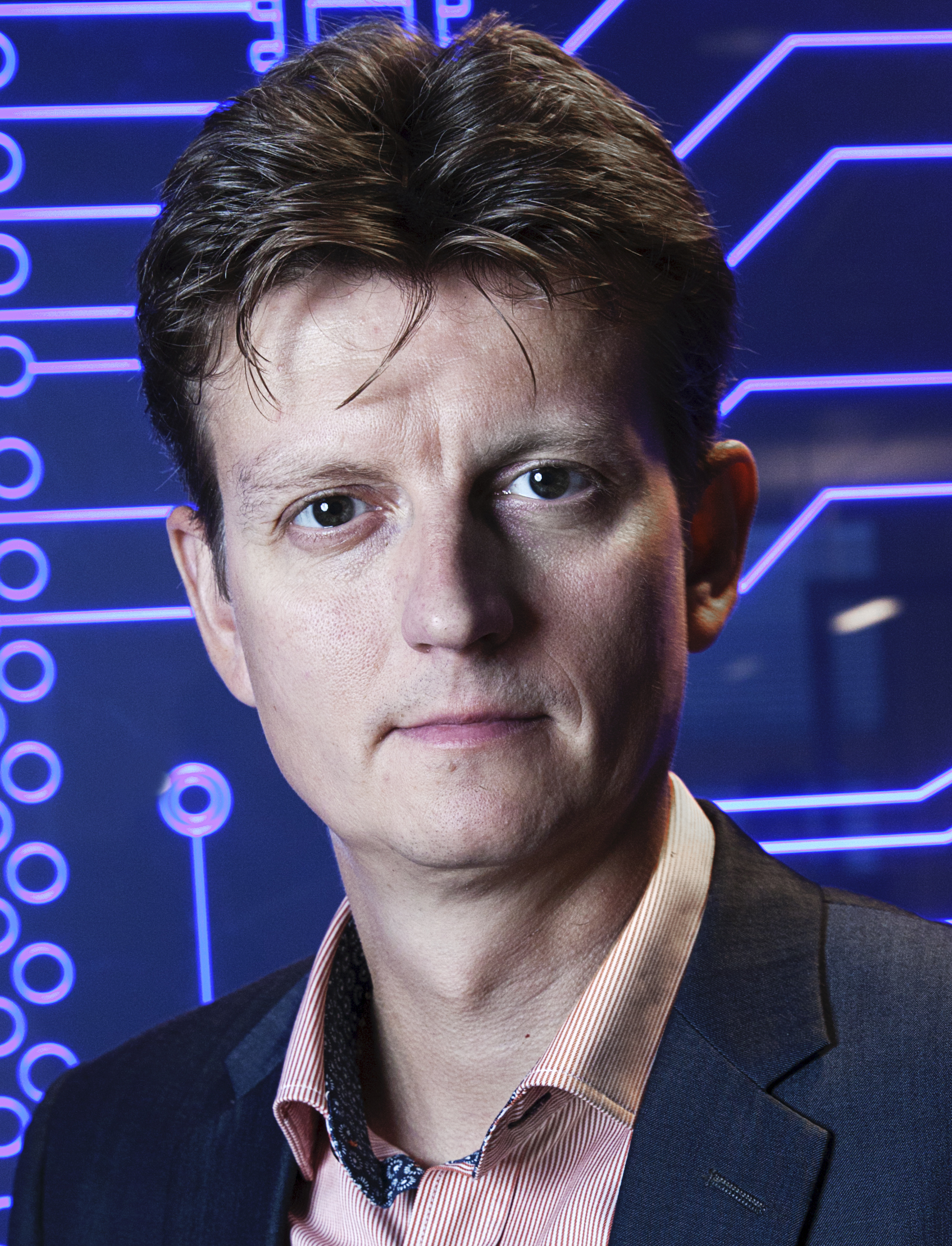}}]{Evgeny Osipov}
received the Ph.D. degree in computer science from the University of Basel, Switzerland, in 2005. 
He is currently a Full Professor in dependable communication and computation systems with the Department of Computer Science and Electrical Engineering, Lule\r{a} University of Technology, Lule\r{a}, Sweden. 

His research interests are in novel  computational models for unconventional computer architectures. 
His current research focuses on hyperdimensional computing and vector symbolic architectures.
\end{IEEEbiography}

\begin{IEEEbiography}[{\includegraphics[width=1in,height=1.25in,clip,keepaspectratio]{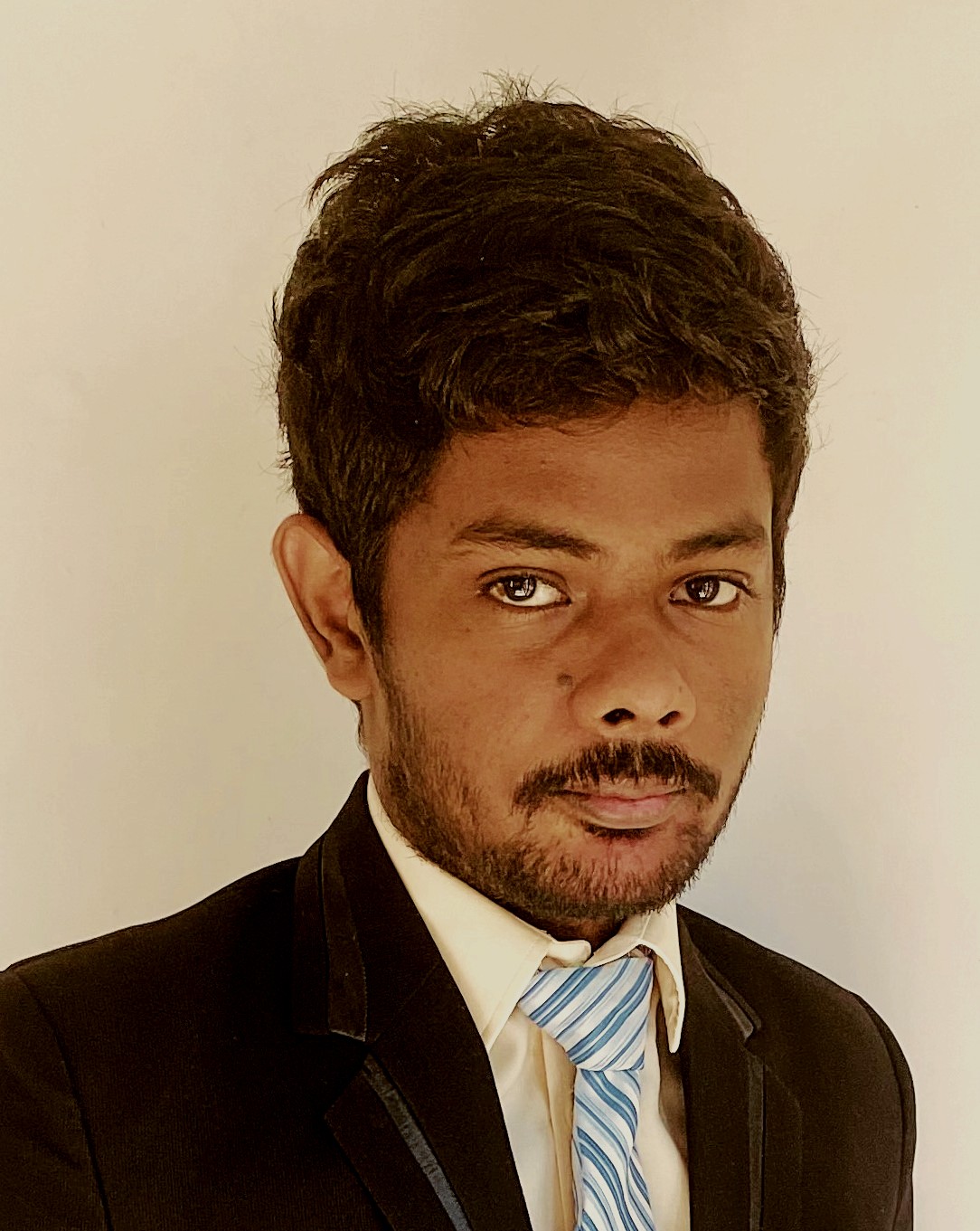}}]{Sachin Kahawala}
received the B.S. degree (Hons.) in Computer Science from the University of Moratuwa, Sri Lanka. He is currently pursuing a Ph.D. in Artificial Intelligence in the Centre for Data Analytics and Cognition at La Trobe University, Australia. His research interests include  hyperdimensional computing, neuromorphic computing, cognitive computing, graph neural networks, energy AI and computer vision.

\end{IEEEbiography}

\begin{IEEEbiography}[{\includegraphics[width=1in,height=1.25in,clip,keepaspectratio]{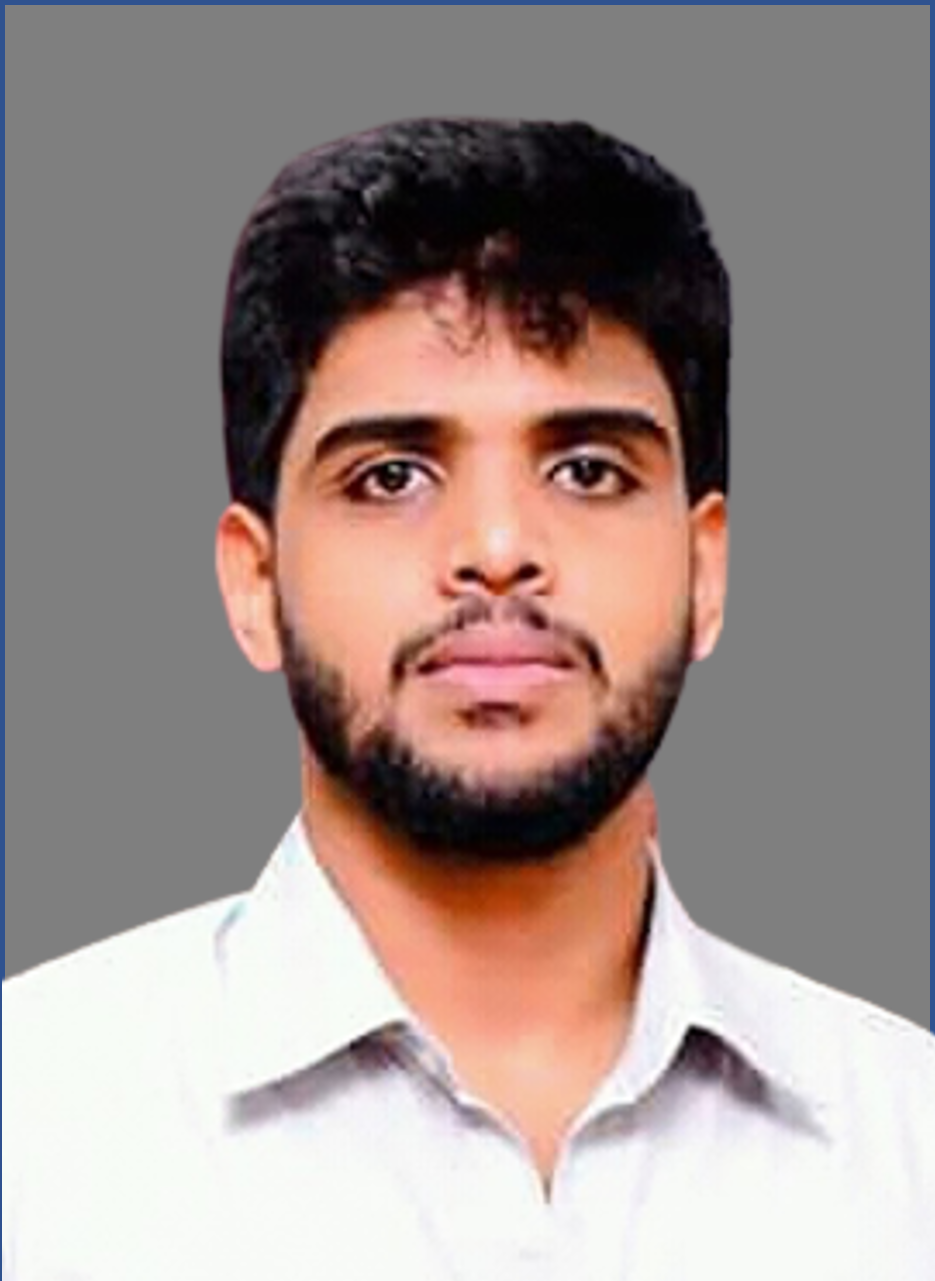}}]{Dilantha Haputhanthri}
completed a B.S. degree (Hons.) in Computer Science and a M.S. degree (Research) in Data Science in the Department of Computer Science and Engineering, University of Moratuwa, Sri Lanka. He is currently pursuing a Ph.D. in Artificial Intelligence in the Centre for Data Analytics and Cognition at La Trobe University, Australia. His research interests include sparse representations, hyperdimensional computing, cognitive computing, and deep learning.

\end{IEEEbiography}

\begin{IEEEbiography}[{\includegraphics[width=1in,height=1.25in,clip,keepaspectratio]{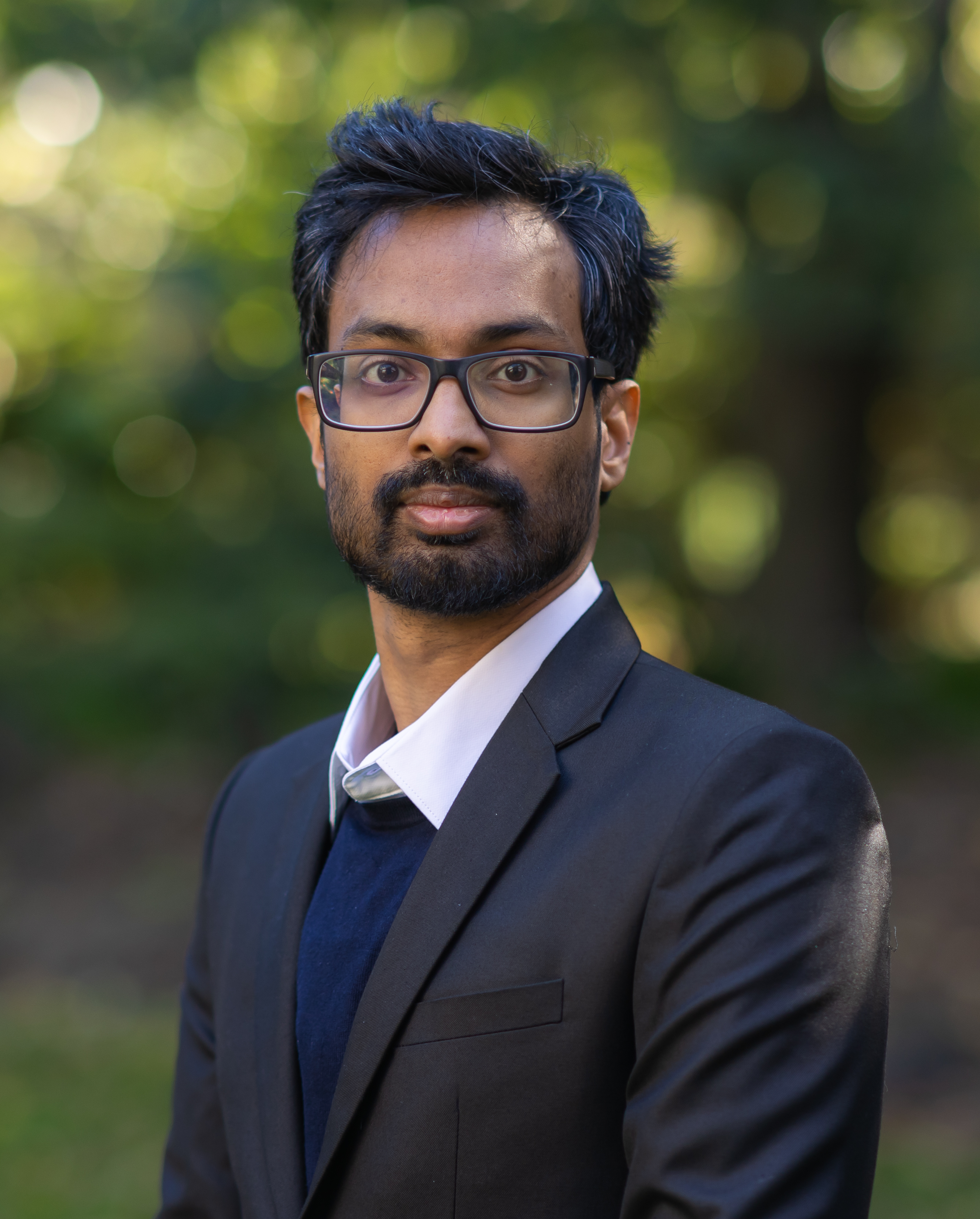}}]{Thimal Kempitiya}
received the B.S. degree (Hons.) from the Department of Computer Science and Engineering, University of Moratuwa, Sri Lanka. He is currently pursuing a Ph.D. in Artificial Intelligence in the Centre for Data Analytics and Cognition, at La Trobe University, Australia. Prior to commencing his Ph.D. degree, he worked as a Technical Lead in a Virtual Reality Healthtech start-up. His research interests include cognitive computing, autonomous systems, digital twins, and data stream mining.

\end{IEEEbiography}

\begin{IEEEbiography}[{\includegraphics[width=1in,height=1.25in,clip,keepaspectratio]{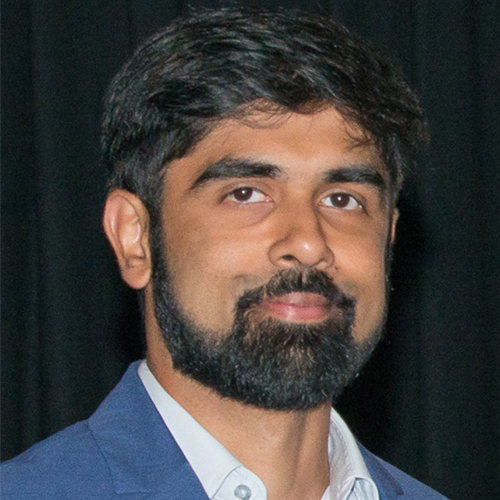}}]{Daswin De Silva}
(Senior Member, IEEE) received the Ph.D. degree in AI from Monash University, Melbourne, Australia. He is the Deputy Director and Course Architect of the Centre for Data Analytics and Cognition, La Trobe University, Australia. His research interests include Data Analytics, Artificial Intelligence (AI), Automation and the Ethics of AI, Data and Machines; with specific focus on autonomous learning, cumulative learning, incremental learning, active perception, information fusion, cognitive computing, NLP, deep emotions, psycholinguistics, and applications in industrial informatics, health, transport, energy, and smart cities. He is an Associate Editor of the following journals, IEEE Transactions on Industrial Informatics, PLOS One, IEEE Open Journal of the Industrial Electronics Society and Springer Discover AI. He was the Lead General Chair of the 15th IEEE International Conference on Human System Interaction - HSI 2022, Melbourne, Australia.

\end{IEEEbiography}

\begin{IEEEbiography}[{\includegraphics[width=1in,height=1.25in,clip,keepaspectratio]{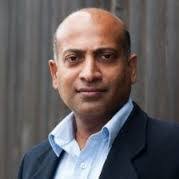}}]{Damminda Alahakoon}
(Member, IEEE) received the Ph.D. degree in AI from Monash University, Melbourne, Australia. He is currently a Full Professor and the Founding Director of the Centre for Data Analytics and Cognition,
La Trobe University, Melbourne, Australia. He has made significant contributions with international impact toward the advancement of AI
through academic research, applied research, research supervision, industry engagement, curriculum development, and teaching. He has published over 100 research articles; theoretical research in self-structuring AI, human-centric AI, cognitive computing, deep learning, optimization, applied AI research in industrial informatics, smart cities, robotics, intelligent transport, digital health, energy, sport science, and education.

\end{IEEEbiography}

\begin{IEEEbiography}[{\includegraphics[width=1in,height=1.25in,clip,keepaspectratio]{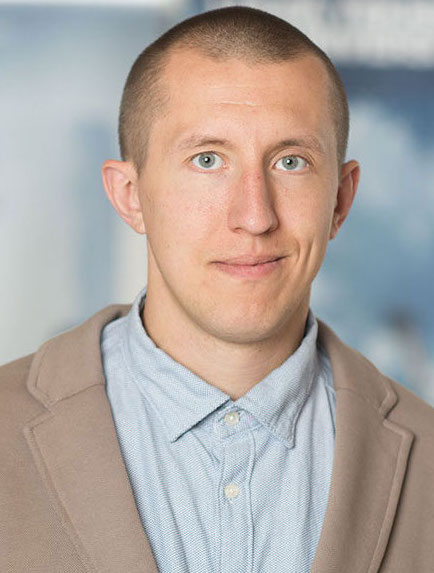}}]{Denis  Kleyko}
(Member, IEEE) received the Ph.D. degree in computer science from the Lule\r{a} University of Technology, Lule\r{a}, Sweden, in 2018.
He is currently a Post-Doctoral Researcher on a joint appointment between the Redwood Center for Theoretical Neuroscience at the University of California at Berkeley, CA, USA and the Intelligent Systems Lab, Research Institutes of Sweden, Kista, Sweden.
His current research interests include machine learning, reservoir computing, and vector symbolic architectures/hyperdimensional computing.  
\end{IEEEbiography}

\end{document}